\documentclass[journal,article,submit,pdftex,moreauthors]{Definitions/mdpi} 
\preto{\abstractkeywords}{\nolinenumbers} 
\usepackage{graphics}
\usepackage{epstopdf}
\usepackage{verbatim}

\firstpage{1} 
\pubvolume{1}
\issuenum{1}
\articlenumber{0}
\pubyear{2022}
\copyrightyear{2022}
\datereceived{} 
\dateaccepted{} 
\datepublished{} 
\hreflink{https://doi.org/} 
\newtheorem{thm}{\bf Theorem}
\newtheorem{dfn}{\bf Definition}
\newtheorem{lem}{\bf Lemma}
\newtheorem{cor}{\bf Corollary}

\Title{Neural Information Squeezer for Causal Emergence}

\TitleCitation{Neural Information Squeezer}

\Author{Jiang  Zhang $^{1,2,*}$, Kaiwei Liu $^{1,\dagger}$}

\AuthorNames{Jiang Zhang and Kaiwei Liu}

\AuthorCitation{Zhang, J.; Liu, K.}

\address{%
$^{1}$ \quad School of Systems Sciences, Beijing Normal University\\
$^{2}$ \quad Swarma Research}

\corres{Correspondence: zhangjiang@bnu.edu.cn}

\abstract{Conventional studies of causal emergence have revealed that stronger causality can be obtained on the macro-level than the micro-level of the same Markovian dynamical systems if an appropriate coarse-graining strategy has been conducted on the micro-states. However, identifying this emergent causality from data is still a hard problem that has not been solved because the appropriate coarse-graining strategy can not be found easily. This paper proposes a general machine learning framework called Neural Information Squeezer to automatically extract the effective coarse-graining strategy and the macro-level dynamics, as well as identify causal emergence directly from time series data. By using invertible neural network, we can decompose any coarse-graining strategy into two separate procedures: information conversion and information discarding. In this way, we can not only exactly control the width of the information channel, but also can derive some important properties analytically. We also show how our framework can extract the coarse-graining functions and the dynamics on different levels, as well as identify causal emergence from the data on several exampled systems.}


\begin{document}
\setcounter{section}{-1} 

\section{Introduction}

Emergence, as one of the most important concepts in complex systems, describes the phenomenon that some overall properties of a system cannot be reduced to the parts\cite{Holland1999Emergence,Bedau1997Emergence}. Causality, as another significant concept, characterises the connection between cause and effect events through time\cite{Pearl2009Causality,Granger1969} for a dynamical system. As pointed out by Hoel et al.\cite{Hoel2013causal_emergence,Hoel2019map}, causality could be emergent, which means that the events of a system on the macro level may have stronger causal connections than the micro level, where the strength of causality could be measured by effective information(EI)\cite{Hoel2013causal_emergence,Tononi2003}. This theoretical framework of causal emergence provides us a new way to understand emergence and other important conceptions in a quantitative way\cite{Varley2021,Chvykov2021,Rosas2020}.

Although many concrete examples of causal emergence across different temporal and spatial scales have been shown in \cite{Hoel2013causal_emergence}, a method to identify causal emergence merely from data is still lack\cite{Rosas2020}. One of the difficulty is how to search all possible coarse-graining strategies (functions, mappings), on which the causal emergence can be shown\cite{Rosas2020}, in a systematic way. On a networked complex system, a coarse-graining strategy includes the way of grouping nodes and the method of mapping the micro-states within a group to a macro-state\cite{Klein2020}. The existing methods solve the problem by fixing the mapping function of states and searching all the grouping methods by heuristic optimization algorithms\cite{Hoel2013causal_emergence,Klein2020}. However, there is no reason why some strategies of state coarse-graining are preferred but not others. Therefore, we should search on the space of all possible coarse-graining strategies such that the most informative dynamics can be identified. Nevertheless, two difficulties we must confront are too large searching space and unavoidability of the trivial mapping between states of micro- and macro. To show the latter, we consider a possible coarse-graining method that maps all the micro-states to an identical value as the macro-state. In this way, the macroscopic dynamics is only an identical mapping that will have large effective information(EI) measure. However, this can not be called causal emergence because all the information is eliminated by the coarse-graining method itself. Thus, we must find a way to exclude such trivial strategies.

An alternative way to identify causal emergence and even other types of emergence is based on partial information decomposition given by \cite{Rosas2020}. Although this method can avoid the discussion on coarse-graining strategies, time consuming searching on subsets of the system state space is also needed. And this method can not give the explicit coarse-graining strategy and the corresponding macro-dynamics which are useful in practice. Furthermore, another common shortage shared by the two mentioned methods is that an explicit Markov transition matrix for both macro- and micro-dynamics are needed, and the transitional probabilities should be estimated from data. As a result, large bias on rare events can hardly be avoided, particularly for continuous data. 

On the other hand, machine learning methods empowered by neural networks have been developed in recent years, and many cross-disciplinary applications have been made\cite{Silver2017,LeCun2015,Reichstein2019,Senior2020}. Equipped with this method, automated discovery of causal relationships and even dynamics of complex systems in a data driven way becomes possible\cite{Tank2018,Lowe2020,Glymour2019,casadiego2017model,sanchez2018graph,zhang2019general,kipf2018neural,chen2021discovering}. Machine learning and neural networks can also help us to find good coarse-graining strategies\cite{Koch-Janusz2018,Li2018,Hu2020,Hu2020RG-Flow,gokmen2021statistical}. If we treat a coarse-graining mapping as a function from micro-states to macro-states, then we can certainly approximate this function by a parameterized neural network. For example, \cite{Hu2020RG-Flow} and \cite{Li2018} used normalized flow model equipped with invertible neural network to learn how to renormalize a multi-dimensional field (quantum field, images or joint probability distributions), and how to generate the field from Gaussian noise. Therefore, both the coarse-graining strategy and the generative model can be learned from data automatically. 

These techniques can also help us to reveal causality on macro-level from data. Causal representation learning aims to use unsupervised representation learning to extract causal latent variables behind the observational data\cite{Chalupka2017,Scholkopf2021}. The encoding process from the original data to the latent causal variables can be understood as a kind of coarse-graining. This shows the similarity between causal emergence identification and causal representation learning, however, their basic objectives are different. Causal representation learning aims to extract the causality hidden in data, whereas, causal emergence identification aims to find a good strategy of coarse-graining to reduce the given micro-level dynamics. Furthermore, introducing multi-scale modeling and coarse-graining operations into causal models brings some new theoretical problems\citep{Iwasaki1994, Rubenstein2017,Beckers2020}. For example, \citep{Rubenstein2017,Beckers2020} discuss the basic requirements of the model abstraction (coarse-graining). However, these studies only care about static random variables and structural causal models but not markovian dynamics. 

In this paper, we formulate the problem of causal emergence identification as a maximization problem of the effective information (EI) for the macro-dynamics under the constraint of precise prediction of micro-dynamics. We then propose a general machine learning framework called Neural Information Squeezer (NIS) to solve the problem. By using invertible neural network to model the coarse-graining strategy, we can decompose any mapping from $\mathcal{R}^{p}$ to $\mathcal{R}^q$ ($q\leq p$) into a series of information conversions invertible processes and information discarding processes. In this way, the framework can not only allow us to control information conversion and discarding in a precise way but also enable us to mathematically analyze the whole framework in theory. We prove a series of mathematical theorems to reveal the properties of NIS. At last, we show how NIS can learn effective coarse-graining strategies and macro-state dynamics numerically on a set of examples. 

\section{Basic Notions and Problems Formulation}
First, we will formulate our problems under a general setting, and layout our framework to solve the problems.

\subsection{Background}

Suppose the dynamics of the complex system that we consider can be described by a set of differential equations.

\begin{equation}
\label{eq:dynamical_model}
    \frac{d\mathbf{x}}{dt}=g\left(\mathbf{x}(t),\xi\right),
\end{equation}
where $\textbf{x}(t)\in \mathcal{R}^p$ is the state of the system and $p\in \mathcal{Z}^+$ is a positive integer, $\xi$ is a random variable of noise. Normally, micro-dynamic $g$ is always markovian which means it could be also modeled as a conditional probability $Pr(\mathbf{x}(t+dt)|\mathbf{x}(t))$ equivalently.

However, we can not directly obtain the evolution of the system but the discrete samples of the states, and we define these states as micro-states.

\begin{dfn}
\label{dfn.micro-states}
(Micro-states): Each sample of the state of the dynamical system (Equation \ref{eq:dynamical_model}) $\mathbf{x}_t$ is called a micro-state at time step $t$. And the multi-variate time series $\mathbf{x}_1,\mathbf{x}_2,\cdot\cdot\cdot, \mathbf{x}_T$ which are sampled with equal intervals and a finite time step $T$, forms a micro-state time series.
\end{dfn}

We always want to reconstruct $g$ according to the observable micro-states. However, an informative dynamical mechanism $g$ with strong causal connections is always hard to be reconstructed from the micro-states when noise $\xi$ is strong. While we can ignore some information in the micro-state data and convert it into macro-state time series. In this way, we may reconstruct a macro-dynamic with stronger causality to describe the evolution of the system. This is the basic idea behind causal emergence\cite{Hoel2013causal_emergence,Hoel2019map}. We formalize the information ignoring process as a coarse-graining strategy(or mapping, method).

\begin{dfn}
\label{dfn.coarse-graining}($q$ dimensional coarse-graining strategy): Suppose the dimension of the macro-states is $0<q<p \in \mathcal{Z}^+$, a $q$ dimensional coarse-graining strategy is a function to map the micro-state $\mathbf{x}_t\in \mathcal{R}^p$ to a macro-state $\mathbf{y}_t\in \mathcal{R}^q$. The coarse-graining is denoted as $\phi_q$. 
\end{dfn}

After coarse-graining, we obtain a new time series data of macro-states denoted by $\mathbf{y}_1=\phi_q(\mathbf{x}_1), \mathbf{y}_2=\phi_q(\mathbf{x}_2),\cdot\cdot\cdot, \mathbf{y}_T=\phi_q(\mathbf{x}_T)$. We then try to find another dynamical model(or a markov chain) $\hat{f}_{\phi_q}$ to describe the evolution of $\mathbf{y}_t$:

\begin{dfn}
\label{dfn.macro-dynamics}(macro-state dynamics): A macro-state dynamics is a set of differential equations
\begin{equation}
\label{eq:macro_dynamics}
    \frac{d \mathbf{y}}{dt}=\hat{f}_{\phi_q}\left(\mathbf{y},\xi'\right),
\end{equation}
such that the solution of Equation \ref{eq:macro_dynamics}, $\mathbf{y}(t)$ is closed to the macro-states $\mathbf{y}_t$ as possible as we can. That is we try to minimize$||\mathbf{y}_t-\mathbf{y}(t)||$ for any $t=1,2,...,T$, where $||\cdot||$ is any norm for vectors. Where $\xi'$ is the noise in the macro-state dynamics.
\end{dfn}

However, this formulation can not reject some trivial strategies. For example, suppose a $q=1$ dimensional $\phi_q$ is defined as $\phi_q(\mathbf{x}_t)=1$ for $\forall \mathbf{y}_t\in \mathcal{R}^p$. Thus, the corresponding macro-dynamic is simply $d\mathbf{y}/dt=0$ and $\mathbf{y}(0)=1$. But this is meaningless because the macro-state dynamic is trivial and coarse-graining mapping is too arbitrary.

Therefore, we must set limitations on coarse-graining strategies and macro-dynamics so that such trivial strategies and dynamics could be avoided. 

\subsection{Effective Coarse-graining Strategy and Macro-dynamics}

We define an effective coarse-graining strategy to be a compressed map such that the macro-states may preserve the information of micro-states as much as it can. Formally,

\begin{dfn}
\label{dfn.effective}
(Effective $q$ coarse-graining strategy and macro-dynamcis): A $q$ coarse-graining strategy $\phi_q:\mathcal{R}^p\rightarrow \mathcal{R}^q$ is effective if there exists a function $\phi_q^{\dagger}:\mathcal{R}^q \rightarrow \mathcal{R}^p$, such that the following inequality holds for a given small real number $\epsilon$:
\begin{equation}
\label{eq:effectiveness}
    ||\phi_q^{\dagger}(\mathbf{y}(t))-\mathbf{x}_t||<\epsilon,
\end{equation}
and the derived macro-dynamic $\hat{f}_{\phi_q}$ is also effective. Where, $\mathbf{y}(t)$ is the solution of equation \ref{eq:macro_dynamics}, that is:

\begin{equation}
\label{eq:solution}
    \mathbf{y}(t)=\phi_q(\mathbf{x}_{t-1})+\int_{t-1}^t\hat{f}_{\phi_q}(\mathbf{y}(\tau),\xi')d\tau
\end{equation}
for all $t=1,2,...,T$. That is, we can reconstruct the micro-state time series by $\phi_q^{\dagger}$ such that the macro-state variables contain the information of micro-states as much as they can. 
\end{dfn}

Notice that this definition is in accordance with the approximate causal model abstraction \cite{Beckers2019causal_abstraction}.

\subsection{Problem Formulation}
Our final objective is to find a most informative macro-dynamic. Therefore, we need to optimize the coarse-graining strategy and the macro-dynamic among all possible effective strategies and dynamics. Therefore, our problem can be formulated as:

\begin{equation}
\label{eq:maximize_ei}
    \max_{\phi_q,\hat{f}_{\phi_q},\phi_q^{\dagger},q} \mathcal{I}(\hat{f}_{\phi_q}),
\end{equation}
under the constraint equations \ref{eq:effectiveness} and \ref{eq:solution}. Where, $\mathcal{I}$ is a measure of effective information, it could be $EI$, $Eff$, or dimension averaged EI which is mainly used in this paper and is denoted as $dEI$(will mention in section \ref{sec.dei}.  $\phi_q$ is an effective coarse-graining strategy, and $\hat{f}_{\phi_q}$ is an effective macro-dynamic.

\section{Methods}
The problem(equation \ref{eq:maximize_ei} and \ref{eq:effectiveness}) is hard to solve because the objects that we will optimize are functions: $\phi_q, \hat{f}_{\phi_q},\phi_q^{\dagger}$ but not numbers. Thus, we use neural networks to parameterize the functions and convert the function optimization problem into a parameter optimization problem.

\subsection{Neural Information Squeezer Model}
\begin{figure}
    \centering
    \includegraphics[scale=0.65]{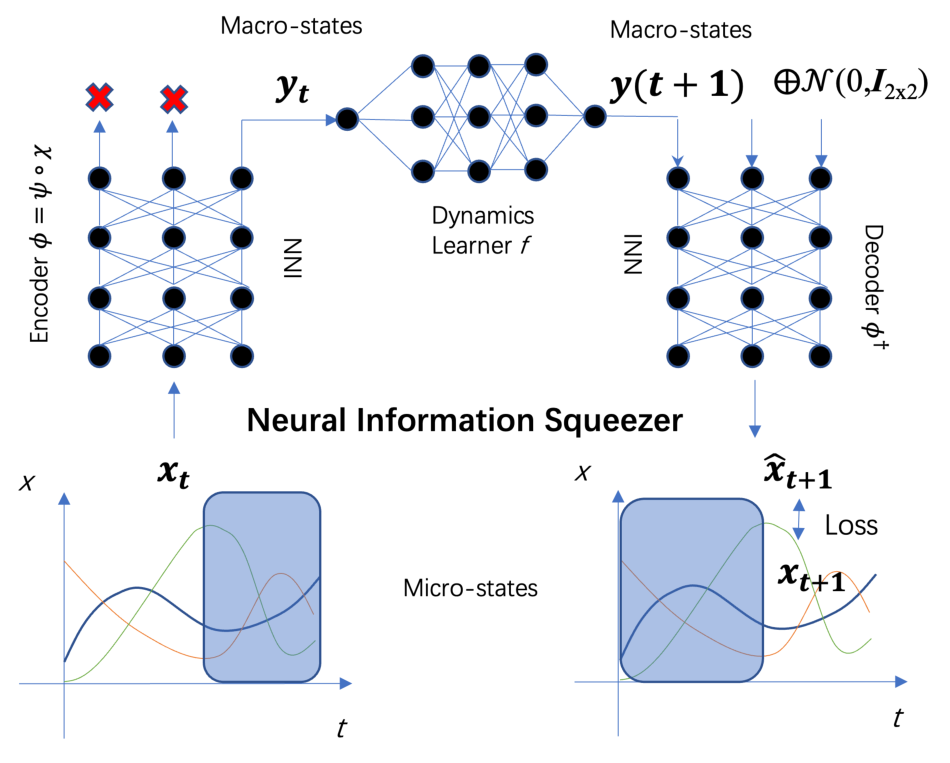}
    \caption{The Workflow and The Framework of the Neural Information Squeezer}
\label{fig:architecture}
\end{figure}

We propose a new machine learning framework called neural information squeezer (NIS) which is based on invertible neural network to solve the problem(equation \ref{eq:maximize_ei}). NIS is composed of three components: encoder, dynamics learner, and decoder. They are represented by neural networks $\psi_{\alpha}$, $f_{\beta}$, and $\psi_{\alpha}^{-1}$ with the parameters $\alpha,\beta$, and $\alpha$ respectively. The entire framework is shown in Figure \ref{fig:architecture}. Next, we will describe each module separately.

\subsubsection{Encoder}

To be noticed, $\psi_{\alpha}$ is an invertible neural network(INN), therefore $\psi$ and $\psi^{-1}$ share the parameters $\alpha$. However, invertible function has no information loss, we must introduce a new operator, projection.

\begin{dfn}
(Projection operator): A projection operator $\chi_{p,q}$ is a function from $\mathcal{R}^p$ to $\mathcal{R}^q$, such that:
\begin{equation}
    \chi_{p,q}(\mathbf{x}_q\bigoplus\mathbf{x}_{p-q})=\mathbf{x}_q,
\end{equation}
where, $\bigoplus$ is the operation of vector concatenation, and $\mathbf{x}_q\in\mathcal{R}^q,\mathbf{x}_{p-q}\in\mathcal{R}^{p-q}$. Sometimes, we abbreviate $\chi_{p,q}$ as $\chi_q$ if there is no ambiguity.
\end{dfn}

Thus, the encoder($\phi$) maps the micro-state $\mathbf{x}_t$ to the macro-state $\mathbf{y}_{t}$, and this mapping can be separated into two steps. That is,

\begin{equation}
    \label{eq:encoding}
    \phi_q = \chi_q \circ \psi_{\alpha},
\end{equation}
where $\circ$ represents the operation of function composition.

The first step is a bijective(invertible) mapping $\psi_{\alpha}:\mathcal{R}^p\rightarrow \mathcal{R}^p$ from $\mathbf{x}_t\in \mathcal{R}^p$ to $\mathbf{x}'_t\in \mathcal{R}^p$ without information lose and is realized by an invertible neural network, the second step is to project the resulting vector to $q$ dimension by mapping $\mathbf{x}'_t\in \mathcal{R}^p$ into $\mathbf{y}_t\in \mathcal{R}^q$ by discarding the information on $p-q$ dimension.

\begin{figure}
\centering
\includegraphics[scale=0.6]{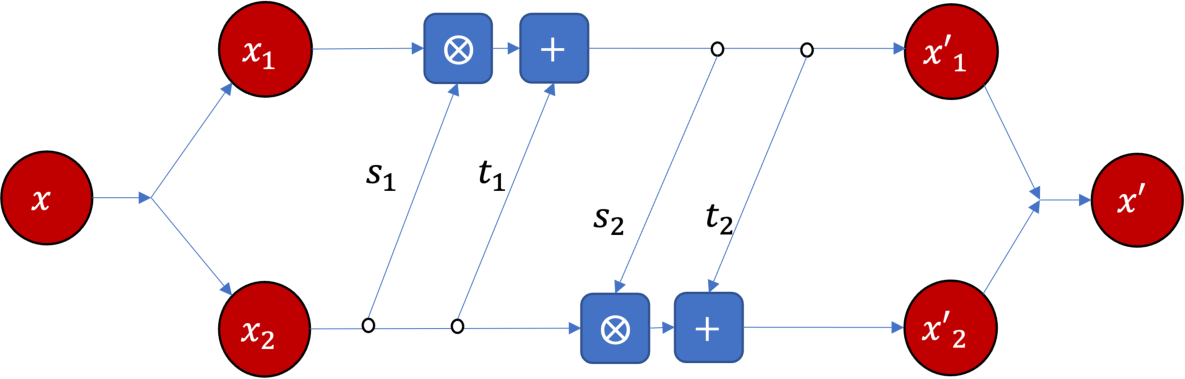}
\caption{The RealNVP neural network implementation of the basic module of the bijector $\psi$. Where, $s_1, s_2$ and $t_1, t_2$ are all feed-forward neural networks with three layers, 64 hidden neurons, and ReLU active function. $s_i$s and $t_i$s share parameters, respectively. $\bigotimes$ and $+$ represent element-wised product and addition, respectively. $\mathbf{x}=\mathbf{x}_1\bigoplus \mathbf{x}_2$ and $\mathbf{x'}=\mathbf{x'}_1\bigoplus \mathbf{x'}_2$.}
\label{fig:realvnp}
\end{figure}

There are several ways to realize an invertible neural network \cite{Teshima2020inn,Teshima2020ode}. While, we select RealNVP module \cite{dinh2016density} as shown in Figure \ref{fig:realvnp} to concretely implement the invertible computation.

In the module, the input vector $\mathbf{x}$ can be separated into two parts, both vectors will be scaled, translated and merged again. The magnitude of the scaling and translation operations will be adjusted by the corresponding feed-forward neural networks. $s_1,s_2$ are the same neural networks shared parameters for scaling, $\bigotimes$ represents element-wised product. And $t_1,t_2$ are the neural networks shared parameters for translation. In this way, an invertible computation from $\mathbf{x}$ to $\mathbf{y}$ can be realized. The same module can be repeated for multiple times (three times in this paper) to realize complex invertible computation, the details can be referred to Appendix \ref{sec.realnvp}. 

The reasons why we use invertible neural network are: 1) INN can reduce the complexity of the model by multiplexing the structure and the parameters in the encoder to the decoder because we can simply reverse the running direction of the encoder to implement decoding; 2) The encoder equipped with INN can separate out the information conversion process and information discarding process; 3) This enables us to do mathematical analysis on the whole framework, and several theorems reflecting the basic properties can be proved.

\subsection{Decoder}
The decoder converts the predicted macro-state of the next time step $\mathbf{y}(t+1)$ into the prediction of the micro-state at the next time step $\hat{\mathbf{x}}_{t+1}$. In our framework, because the coarse-graining strategy $\phi_q$ can be decomposed as a bijector $\psi_{\alpha}$ and a projector $\chi_q$, we can simply reverse $\psi_{\alpha}$ to become $\psi^{-1}_{\alpha}$ as the decoder. However, because the dimension of the macro-state is $q$ and the input dimension of $\psi_{\alpha}$ is $p>q$, we need to fill the remaining $p-q$ dimensions by a $p-q$ dimensional Gaussian random vector. That is, for any $\phi_q$, the decoding mapping can be defined as:
\begin{equation}
    \phi_q^{\dagger}=\psi_{\alpha}^{-1}\circ\chi_q^{\dagger},
\end{equation}
where $\psi_{\alpha}^{-1}$ is the inverse function of $\psi_{\alpha}$, and $\chi_q^{\dagger}: \mathcal{R}^q\rightarrow \mathcal{R}^p$ is a function defined as follow: for any $\mathbf{x}_q\in\mathcal{R}^p$
\begin{equation}
    \chi_q^{\dagger}(\mathbf{x}_q)=\mathbf{x}_q\bigoplus \mathbf{z}_{p-q},
\end{equation}
where $\mathbf{z}_{p-q}\sim \mathcal{N}(0,\mathcal{I}_{p-q})$ is a random Gaussian noise with $p-q$ dimension, and $\mathcal{I}_{p-q}$ is an identity matrix with the same dimension. That is, we can generate a micro-state by composing $\mathbf{x}_q$ and a random sample $\mathbf{z}_{p-q}$ from a $p-q$ dimensional standard normal distribution. 

According to the point view of \cite{Li2018,Hu2020RG-Flow}, the decoder can be regarded as a generative model of the conditional probability $Pr(\hat{\mathbf{x}}_{t+1}|\mathbf{y}(t+1))$, and the encoder just performs a renormalization process.

\subsubsection{Dynamics Learner}
The dynamics learner $f_{\beta}$ is a common feed-forward neural network with parameters $\beta$, it will learn the effective markov dynamic on the macro-level. Concretely, we at first use $f_{\beta}$ to replace $\hat{f}_{\phi_q}$ in equation \ref{eq:macro_dynamics}, and second we use Euler method with $dt=1$ to solve the Equation \ref{eq:macro_dynamics}, and suppose the noise is a additive Gaussian(or Laplacian)\cite{Kingma2013}, therefore we can reduce Equation \ref{eq:solution} as:
\begin{equation}
    \mathbf{y}(t+1)=\mathbf{y}_t +\int_t^{t+1}f_{\beta}(\mathbf{y}(\tau),\xi')d\tau\approx \mathbf{y}_t+f_{\beta}(\mathbf{y}_t)+\xi'
\end{equation}
where $\xi'\sim \mathcal{N}(0,\Sigma)$ or $Laplacian(0,\Sigma)$, $\Sigma=diag(\sigma_1^2,\sigma_2^2,\cdot\cdot\cdot,\sigma_q^2)$ is the covariance matrix, and $\sigma_i$ is the standard deviation in the $i$th dimension which could be learned or fixed. Thus, the transitional probability of this dynamics can be written as
\begin{equation}
    P(\mathbf{y}(t+1)|\mathbf{y}_t)=\mathcal{D}(\mu(\mathbf{y}_t),\Sigma),
\end{equation}
where $\mathcal{D}$ represents the PDF of Gaussian distribution or Laplace distribution, $\mu(\mathbf{y}_t)\equiv \mathbf{y}_t+f_{\beta}(\mathbf{y}_t)$ is the mean vector of the distribution.

By training the dynamics learner in an end-to-end manner, we can avoid estimating the markov transitional probabilities from the data to reduce biases because neural networks always have much better ability to fit the data and generalize to unseen cases.

\subsection{Two stage optimization}
Although the functions that will be optimized have been parameterized by neural networks, Equation \ref{eq:maximize_ei} is still hard to be optimized directly because the objective function and the constraint condition must be combined together to be considered and $q$ as a hyper-parameter can affect the structure of neural networks. Thus, In this paper, we propose a two-stage optimization method. In the first stage, we fix the hyper-parameter $q$ and optimize the difference between the predicted micro-state and the observed data $|\phi_q^{\dagger}(\mathbf{y}(t))-\mathbf{x}_t|$, that is Equation \ref{eq:effectiveness}, to let the coarse-graining strategy $\phi_q$ and macro-dynamics $\hat{f}_q$ to be effective. And then, we search for all possible $q$ values to find the optimal one such that $\mathcal{I}$ can be maximized.

\subsubsection{Stage 1: training a predictor}
In the first stage, we can use likelihood maximization and stochastic gradient descend techniques to obtain the effective $q$ coarse-graining strategy and the effective predictor of the macro-state dynamics. The objective function is defined on the likelihood of micro-state prediction. 

We can understand a feed-forward neural network as a machine to model a conditional probability with Gaussian or Laplacian distribution\cite{Kingma2013}. Thus, the entire NIS framework can be understood as a model of $P(\hat{\mathbf{x}}_{t+dt}|\mathbf{x}_t)$ with the output $\hat{\mathbf{x}}_{t+1}$ is just the mean value. And the objective function Equation \ref{eq:objectivefunction} is just the log-likelihood or cross-entropy of the observed data under the given form of the distribution.

\begin{equation}
    \label{eq:loglikelihood}
    \mathcal{L}=\sum_t \ln P(\hat{\mathbf{x}}_{t+1}=\mathbf{x}_{t+1}|\mathbf{x}_t),
\end{equation}
where $P(\hat{\mathbf{x}}_{t+1}= \mathbf{x}_{t+1}|\mathbf{x}_t)\equiv\mathcal{N}(\hat{\mathbf{x}}_{t+1}, \Sigma)$ when $l=2$ or $Laplace(\hat{\mathbf{x}}_{t+1}, \Sigma)$ when $l=1$, where $\Sigma$ is the covariance matrix which is always be a diagonal matrix and the magnitude can be calculated as the mean square error for $l=2$ or mean absolute value for $l=1$.

If we take the concrete form of Gaussian or Laplacian distribution into the conditional probability, we will see to maximize the log-likelihood is equivalent to minimize the $l$-norm objective function:

\begin{equation}
    \label{eq:objectivefunction}
    \mathcal{L}=\sum_t||\hat{\mathbf{x}}_{t+1}-\mathbf{x}_{t+1}||_{l}
\end{equation}
where $l=1$ or $2$. 

Then we can use stochastic gradient descend technique to optimize Equation \ref{eq:objectivefunction}.

\subsubsection{Stage 2: search for the optimal scale}
In the previous step, we can obtain the effective $q$ coarse graining strategy and the macro-state dynamics after a large number of training epochs, but the results are dependent on $q$. 

To select the optimized $q$, we can compare the measure of effective information $\mathcal{I}$ for different $q$ coarse-graining macro-dynamics. Because the parameter $q$ only has one dimension, and its value range is also limited ($0<q<p$), we can simply iterate all $q$ to find out the optimal $q^*$ and the optimal effective strategy.

\subsubsection{About Effective Information}
\label{sec.dei}
In the second stage, to compare coarse-graining strategies and macro-dynamics, we need to compute the important indicator: effective information (EI), however, the conventional computations of EIs are all for discrete markov dynamics in most of previous works\cite{Hoel2013causal_emergence,Hoel2019map}, and we may confront difficulties when we apply EI on continuous dynamics\cite{Chvykov2021}. 

First, the conventional methods on mutual information computation for discrete variables cannot be used here, new methods for continuous variables and mappings especially for high dimensional space must be invented. To solve the problem, we treat the mapping of the dynamics learner neural network as an conditional Gaussian distribution, thereafter, we can calculate EI for this Gaussian distribution. Concretely, we have the following theorem:

\begin{thm}
\label{thm.ei_gauss}
(EI for feed-forward neural networks) In general, if the input of a neural network is $X=(x_1,x_2,\cdot\cdot\cdot,x_n)\in [-L,L]^n$, which means $X$ is defined on a hyper-cube with size $L$, where $L$ is a very large integer. The output is $Y=(y_1,y_2,\cdot\cdot\cdot,y_m)$, and $Y=\mu(X)$. Here $\mu$ is the deterministic mapping implemented by the neural network: $\mu: \mathcal{R}^n\rightarrow \mathcal{R}^m$, and its Jacobian matrix at $X$ is $\partial_{X'} \mu(X)\equiv \left\{\frac{\partial \mu_i(X')}{\partial X'_j}\left|_{X'=X}\right.\right\}_{nm}$. If the neural network can be regarded as a Gaussian distribution conditional on given $X$:
\begin{equation}
    \label{eq:gaussian}
    p(Y|X)=\frac{1}{\sqrt{(2\pi)^m|\Sigma|}}\exp{\left(-\frac{1}{2}(Y-\mu(X))^T\Sigma^{-1}(Y-\mu(X))\right)}
\end{equation}
where, $\Sigma=diag(\sigma_1^2,\sigma_2^2,\cdot\cdot\cdot,\sigma_m^2)$ is the co-variance matrix, and $\sigma_i$ is the standard deviation of the output $y_i$ which can be estimated by the mean square error of $y_i$, then the effective information (EI) of the neural network can be calculated in the following way:

(i) If there exists $X$ such that $\det(\partial_{X'} \mu(X))\neq 0$, then the effective information (EI) can be calculated as:
\begin{equation}
\begin{aligned}
        EI_L(\mu)=I(do(X\sim U([-L,L]^{n};Y)\approx & -\frac{m+m \ln (2\pi)+ \sum_{i=1}^m\sigma_i^2}{2}\\
        & +n\ln (2L) + \mathbb{E}_{X\sim U([-L,L]^n} \left(\ln |\det(\partial_{X'} \mu(X))|\right).
\end{aligned}
\end{equation}
where, $U([-L,L]^n)$ is the uniform distribution on $[-L,L]^n$, and $|\cdot|$ is absolute value, and $\det$ is determinant. 

(ii) If $\det(\partial_{X'} \mu(X))\equiv 0$ for all $X$, then $EI\approx 0$
\end{thm}

Although Theorem \ref{thm.ei_gauss} can solve the problem of EI computation for continuous variables and functions, new problems must be confronted which are:  1) EI will be affected by the output dimension $m$ easily, this may trouble the comparison of EI for different dimensional dynamics, and 2) EI is dependent on $L$, and will be divergent when $L$ is very large.

To solve the first problem, we define a new indicator which is called dimension averaged effective information or effective information per dimension. Formally,

\begin{dfn}
(Dimension Averaged Effective Information (dEI)): For a dynamic $f$ with $n$ dimensional state space, then the dimension averaged effective information is defined as:
\begin{equation}
    dEI(f) = \frac{EI(f)}{n}
\end{equation}
\end{dfn}

Therefore, if the dynamic $f$ is continuous and can be regarded as a conditional Gaussian distribution, then according to Theorem \ref{thm.ei_gauss}, the dimension averaged EI can be calculated as($m=n$):
\begin{equation}
    \label{eqn.dEI}
    dEI_L(f)=-\frac{1+\ln(2\pi)+\sum_{i=1}^n\sigma_i^2/n}{2}+\ln(2L)+\frac{1}{n}\mathbb{E}_{X\sim U([-L,L]^n} \left(\ln |\det(\partial_{X'} f(X))|\right).
\end{equation}

It is easy to see that all the terms related with dimension $n$ in Equation \ref{eqn.dEI} is eliminated. However, there is still $L$ in the equation which may cause divergent when $L$ is very large.

Therefore, to solve this problem, we can calculate the dimension averaged causal emergence (dCE) to eliminate the influence of $L$. 
\begin{dfn}
\label{dfn.dCE}
(Dimension averaged causal emergence(dCE)): for macro-dynamics $f_M$ with dimension $n_M$ and micro-dynamics $f_m$ with dimension $n_m$, we define dimension averaged causal emergence as:
\begin{equation}
    \label{eq.dCE}
    dCE(f_M,f_m)=dEI(f_M)-dEI(f_m)=\frac{EI(f_M)}{n_M}-\frac{EI(f_m)}{n_m}.
\end{equation}
\end{dfn}

Thus, if the dynamics $f_M$ and $f_m$ are continuous and can be regarded as conditional Gaussian distributions, then according to definition \ref{dfn.dCE} and equation \ref{eqn.dEI}, the dimension averaged causal emergence can be calculated as:
\begin{equation}
    \label{eq.dCE_gauss}
    \begin{aligned}
        dCE(f_M,f_m)=&\left(\frac{1}{n_M}\mathbb{E}_{X_M}\ln|\det \partial_{X_M}f_M|-\frac{1}{n_m}\mathbb{E}_{X_m}\ln|\det \partial_{X_m}f_m|\right)\\
        &-\left(\frac{1}{n_M}\sum_{i=1}^{n_M}\ln\sigma_{i,M}^2-\frac{1}{n_m}\sum_{i=1}^{n_m}\ln\sigma_{i,m}^2\right)
    \end{aligned}
\end{equation}

Therefore, all the effects of dimension $n$ and $L$ have been eliminated in Equation \ref{eq.dCE_gauss}, and the result is only influenced by the relative values of the variances and the logarithmic values of the determinant of the jacobian matrices. In the following numeric computations, we will mainly use Equation \ref{eq.dCE_gauss}. The reason why we not use Eff is also because it contains $L$.
\section{Results}
In this section, we will layout several theoretic properties of NIS at first, then we will apply it on some numeric examples.
\subsection{Theoretical Analysis}
To understand why the neural information squeezer framework can find out the most informative macro-dynamics and how the effective strategy and dynamics change with $q$, we at first layout some major theoretical results through mathematical analysis. Notice that although all of the theorems are about mutual information, these conclusions are also suitable for effective information because all the theoretical results are irrelevant to the distribution of input data.

\subsubsection{Squeezed Information Channel}
\begin{figure}
    \centering
    \includegraphics[scale=0.5]{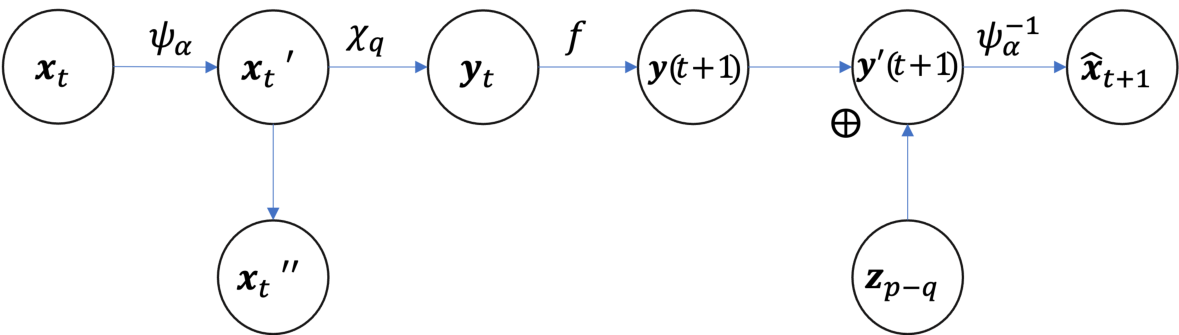}
    \caption{The graphic model of the neural information squeezer as a squeezed information channel}
\label{fig:squeezed_channel1}
\end{figure}

First, we notice that the framework (Figure \ref{fig:architecture}) can be regarded as an information channel as shown in Figure \ref{fig:squeezed_channel1}, and due to the existence of the projection operation, the channel is squeezed in the middle. Therefore, we call that a squeezed information channel(see also Appendix \ref{sec.gaussianei} for formal definition for the sqeezed information channel). 

As proved in Appendix \ref{sec.gaussianei}, we have a theorem for the squeezed information channel:

\begin{thm}
\label{thm.info_bottle}
(Information bottleneck of Information Squeezer): For the squeezed information channel as shown in Figure \ref{fig:squeezed_channel1} and for any bijector $\psi$, projector $\chi_q$, macro-dynamics  $f$, and the random noise $\mathbf{z}_{p-q}\sim \mathcal{N}(0,\mathcal{I}_{p-q})$, we have: 
\begin{equation}
    I(\mathbf{y}_t;\mathbf{y}(t+1))=I(\mathbf{x}_t;\hat{\mathbf{x}}_{t+1}),
\end{equation}
where $\hat{\mathbf{x}}_{t+1}$ is the prediction of NIS, and $\mathbf{y}(t+1)$ follows Equation \ref{eq:macro_dynamics}.
\end{thm}
That is, for any neural network that implements the general framework as shown in Figure \ref{fig:squeezed_channel1}, the mutual information of macro-dynamic $f_{\phi_q}$ is identical to the entire dynamical model, i.e., the mapping from $\mathbf{x}_t$ to $\hat{\mathbf{x}}_{t+1}$ for any time. Theorem \ref{thm.info_bottle} is fundamental for NIS. Actually, the macro-dynamics $f$ is the information bottleneck of the entire channel\cite{shwartz2017opening}. 

\subsubsection{What happens during training}

With Theorem \ref{thm.info_bottle}, we can understand what happens when the neural squeezer framework is trained by data in an intuitive way. 

First, we know as the neural networks are trained, the output of the entire framework $\hat{\mathbf{x}}_{t+1}$ is closed to the real data $\mathbf{x}_{t+1}$ under any given $\mathbf{x}_t$, so do the mutual information, that is the following theorem:

\begin{thm}
\label{thm.training}
(Mutual information of the model will be closed to the data for a well trained framework): If the neural networks in NIS framework are well-trained, then:
\begin{equation}
    I(\hat{\mathbf{x}}_{t+1};\mathbf{x}_t)\simeq I(\mathbf{x}_{t+1};\mathbf{x}_t)
\end{equation}
\end{thm}

The proof is in the Appendix \ref{sec.training}.

Second, we suppose that the mutual information $I(\mathbf{x}_t,\mathbf{x}_{t+1})$ is always large because the time series of micro-states $\mathbf{x}_t$ contains information. Otherwise, we may not be interested in $\{\mathbf{x}_t\}$. Therefore, as the neural network is trained, $I(\mathbf{x}_t;\hat{\mathbf{x}}_{t+1})$ will increase to be closed to $I(\mathbf{x}_t;\mathbf{x}_{t+1})$.

Third, according to Theorem \ref{thm.info_bottle}, $I(\mathbf{y}_t;\mathbf{y}(t+1))=I(\mathbf{x}_t,\hat{\mathbf{x}}_{t+1})$ will also be increased such that it can be closed to $I(\mathbf{x}_t;\mathbf{x}_{t+1})$.

Because the macro-dynamics is the information bottleneck of the entire channel, therefore its information must be increased as training. In the same time, the determinant of the Jacobian of $\psi_{\alpha}$ and the entropy of $\mathbf{y}_t$ will also be increased in a general case. This conclusion is implied in Theorem \ref{thm.lowerbound}.

\begin{thm}
\label{thm.lowerbound}
(Information on bottleneck is the lower bound of the encoder): For the squeezed information channel shown in Figure \ref{fig:squeezed_channel1}, the determinant of the Jacobian matrix of $\psi_{\alpha}$ and the Shannon entropy of $\mathbf{y}_t$ are lower bounded by the information of the entire channel:
\begin{equation}
\label{eq.lowerbound}
    H(\mathbf{x}_t)+\mathbb{E}(\ln|\det(J_{\psi_{\alpha}}(\mathbf{x}_t))|)\geq H(\mathbf{y}_t)+\mathbb{E}(\ln|\det(J_{\psi_{\alpha},\mathbf{y}_t}(\mathbf{x}_t))|)\geq I(\mathbf{x}_t;\hat{\mathbf{x}}_{t+1}),
\end{equation}
where, $H$ is the Shannon entropy measure, $J_{\psi_{\alpha}}(\mathbf{x}_t)$ is the Jacobian matrix of the bijector $\psi_{\alpha}$ at the input $\mathbf{x}_t$, and $J_{\psi_{\alpha},\mathbf{y}_t}(\mathbf{x}_t)$ is the sub-matrix of $J_{\psi_{\alpha}}(\mathbf{x}_t)$ on the projection $\mathbf{y}_t$ of $\mathbf{x}_t'$. 
\end{thm}

The proof is also given in Appendix \ref{sec.lowerbound}.

Because the distribution of $\mathbf{x}_t$ and its Shannon entropy are given, thus, Theorem \ref{thm.lowerbound} states that the expectation of the logrithim of $|\det(J_{\psi_{\alpha}}(\mathbf{x}_t))|$ and the entropy of $\mathbf{y}_t$ must be larger than the information of the entire information channel. 

Therefore, once the initial values of $\mathbb{E}|\det(J_{\psi_{\alpha}}(\mathbf{x}_t))|$ and $\mathbf{y}_t$ are small, as the model is trained, the mutual information of the entire channel increases, the determinant of the Jacobian must also be increased, and the distribution of the macro-state $\mathbf{y}_t$ must be more disperse. But these may not happen if the information $I(\mathbf{x}_t;\hat{\mathbf{x}}_{t+1})$ has been closed to $I(\mathbf{x}_t;\mathbf{x}_{t+1})$ or $\mathbb{E}|\det(J_{\psi_{\alpha}}(\mathbf{x}_t))|$ and $H(\mathbf{y}_t)$  have been already large enough.

\subsubsection{The Effective Information is mainly determined by the Bijector}
The previous analysis is about the mutual information but not the effective information of the macro-dynamic which is the key ingredients about causal emergence. Actually, with the good properties of the squeezed information channel, we can write down an expression of the $EI$ for the macro-dynamic but without the explicit form of it. And, accordingly, we find the major ingredient to determine causal emergence is the bijector $\psi_{\alpha}$.

The proof is detailed in Appendix \ref{sec.ei_calculation}

\begin{thm}
\label{thm.ei_bijection}
(The mathematical expression for effective information of the macro-dynamics): Suppose the probability density of $\mathbf{x}_{t+1}$ under given $\mathbf{x}_t$ can be described by a function $Pr(\mathbf{x}_{t+1}|\mathbf{x}_t)\equiv G(\mathbf{x}_{t+1},\mathbf{x}_t)$, and the Neural Information Squeezer framework is well trained, then the effective information of the macro-dynamics of $f_{\beta}$ can be calculated by:
\begin{equation}
    \label{eq:theorem_ei0}
    EI_L(f_{\beta})=\frac{1}{(2L)^{p}}\cdot\int_{\sigma} \int_{\mathcal{R}^p} G(\mathbf{y},\psi_{\alpha}^{-1}(\mathbf{x}))\ln\frac{(2L)^{p}G(\mathbf{y},\psi_{\alpha}^{-1}(\mathbf{x}))}{\int_{\sigma}G(\mathbf{y},\psi_{\alpha}^{-1}(\mathbf{x}'))d\mathbf{x}'}d\mathbf{y} d\mathbf{x},
\end{equation}
where, $\sigma\equiv [-L,L]^p$ is the integration region for $\mathbf{x}$ and $\mathbf{x}'$.
\end{thm}

\subsubsection{Change with the Scale($q$)}
According to Theorems \ref{thm.info_bottle} and \ref{thm.training}, we have the following corollary \ref{cor.mi}:

\begin{cor}
\label{cor.mi}
(The mutual information of macro-dynamics will not change if the model is well trained): For the well trained NIS model, the Mutual Information of the macro-dynamics $f_{\beta}$ will be irrelevant of all the parameters, including the scale $q$.
\end{cor}

If the neural networks are well-trained, the mutual information on the macro-dynamics will approach to the information in the data $\{\mathbf{x}_t\}$. So no matter how small $q$ is (or how large is the scale), the mutual information of the macro-dynamics $f_{\beta}$ will keep constant.

It seems that the scale $q$ is an irrelevant parameter on causal emergence. However, according to Theorem \ref{thm.harder}, smaller $q$ will lead to the encoder carrying more effective information.

\begin{thm}
\label{thm.harder}
(Narrower is Harder): If the dimension of $\mathbf{x}_t$ is $p$, then for $0<q_1<q_2<p$:
\begin{equation}
\label{eq:theorem5}
    I(\mathbf{x}_t;\hat{\mathbf{x}}_{t+1})\leq I(\mathbf{x}_t;\mathbf{y}_{t}^{q_1})\leq I(\mathbf{x}_t;\mathbf{y}_{t}^{q_2}),
\end{equation}
where $\mathbf{y}_{t}^{q}$ denotes the $q$-dimensional vector $\mathbf{y}_{t}$.
\end{thm}

The mutual information in Theorem \ref{thm.harder} is about the encoder, i.e., the micro-state $\mathbf{x}_t$ and the macro-state $\mathbf{y}_t$ in different dimension $q$. The theorem states that as $q$ decreases, the mutual information of the encoder part must also decrease and more closed to the information limitation $I(\mathbf{x}_t;\hat{\mathbf{x}}_{t+1})\simeq I(\mathbf{x}_t;\mathbf{x}_{t+1})$. Therefore, the entire information channel becomes narrower, the encoder must carry more useful and effective information to transfer to the macro-dynamics. And the prediction becomes harder.

\subsection{Empirical Results}

We test our model on several data sets. All the data is generated by the simulated dynamical models. And the models include continuous dynamics and discrete Markovian dynamics.

\subsubsection{Spring Oscillator with Measurement Noise}
The first experiment to test our model is a simple spring oscillator following the dynamical equations:
\begin{equation}
    \label{eq:spring_oscillator}
        \left\{
        \begin{aligned}
        &dz/dt = v\\
        &dv/dt = -z
        \end{aligned}
        \right.
\end{equation}
where, $z$ and $v$ are position and velocity of the oscillator in one dimension, respectively. The states of the system can be represented as $\mathbf{x}=(z,v)$.

However, we can only observe the state from two sensors with measurement errors. Suppose the observational model is
\begin{equation}
    \label{eq:obervation}
    \left\{
        \begin{aligned}
    \mathbf{\tilde{x}}_1=\mathbf{x}+\zeta\\
    \mathbf{\tilde{x}}_2=\mathbf{x}-\zeta
    \end{aligned}
    \right.
\end{equation}
where, $\zeta\sim \mathcal{N}(0,\sigma)$ is a random number following two dimensional Gaussian distribution, and $\sigma$ is the vector of the standard deviations for position and velocity. In this example, we can understand the states $\mathbf{x}$ as latent macro-states and the measurements $\tilde{\mathbf{x}}_1,\tilde{\mathbf{x}}_2$ are micro-states. What will NIS do is to recover the latent macro-state $\mathbf{x}$ from the measurements.

According to Equation \ref{eq:obervation}, although there is noise to disturb the measurement of the state, it can be easily eliminated by adding the measurements on the two channels together. Therefore, if NIS can discover a macro-state which is the addition of the two measurements, then it can easily obtain the correct dynamics. We sample the data for 10,000 batches (with Euler method and $dt=1$), and in each batch, we randomly generate 100 random initial states and perform one step dynamic to get the state at the next time step. We use these data to train the neural network. To compare, we also use the same data set to train an ordinary feed-forward neural network with the same number of parameters.

\begin{figure}[htbp]
    \begin{minipage}[c]{0.5\textwidth}
        \centering
        \includegraphics[width=1\textwidth]{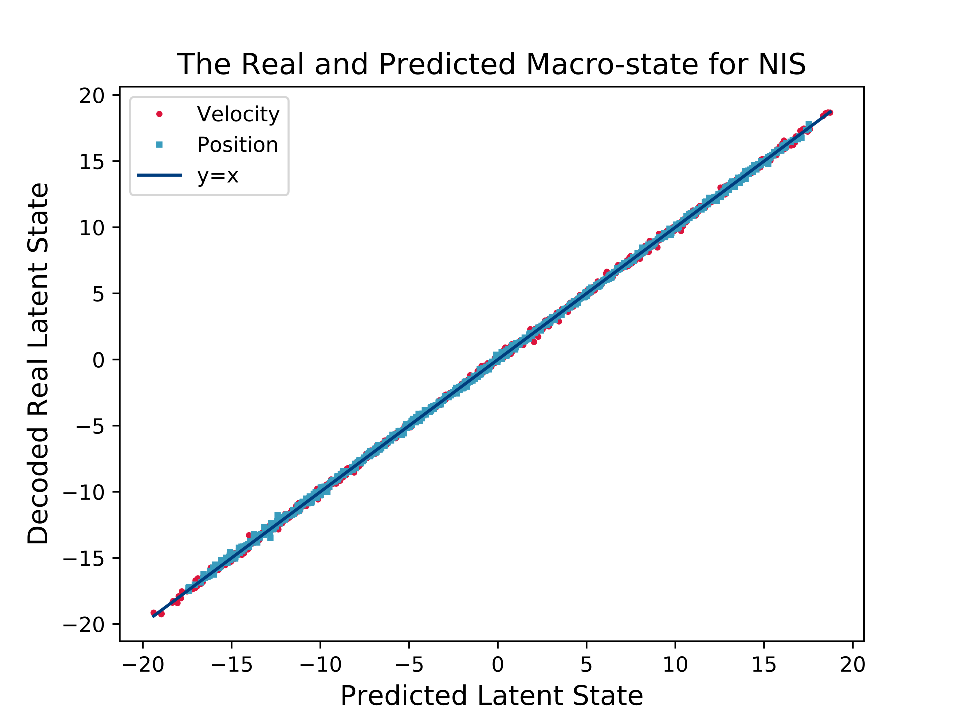}
        \centerline{(a)}
    \end{minipage}
    \begin{minipage}[c]{0.5\textwidth}
        \centering
        \includegraphics[width=1\textwidth]{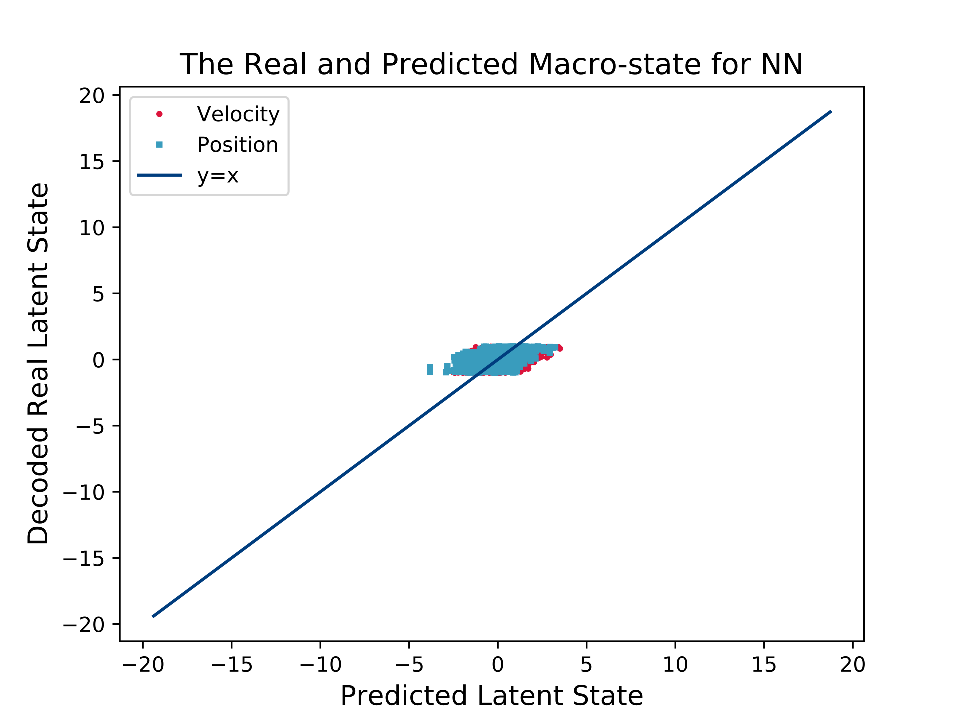}
        \centerline{(b)}
    \end{minipage}
    \begin{minipage}[c]{0.5\textwidth}
        \centering
        \includegraphics[width=1\textwidth]{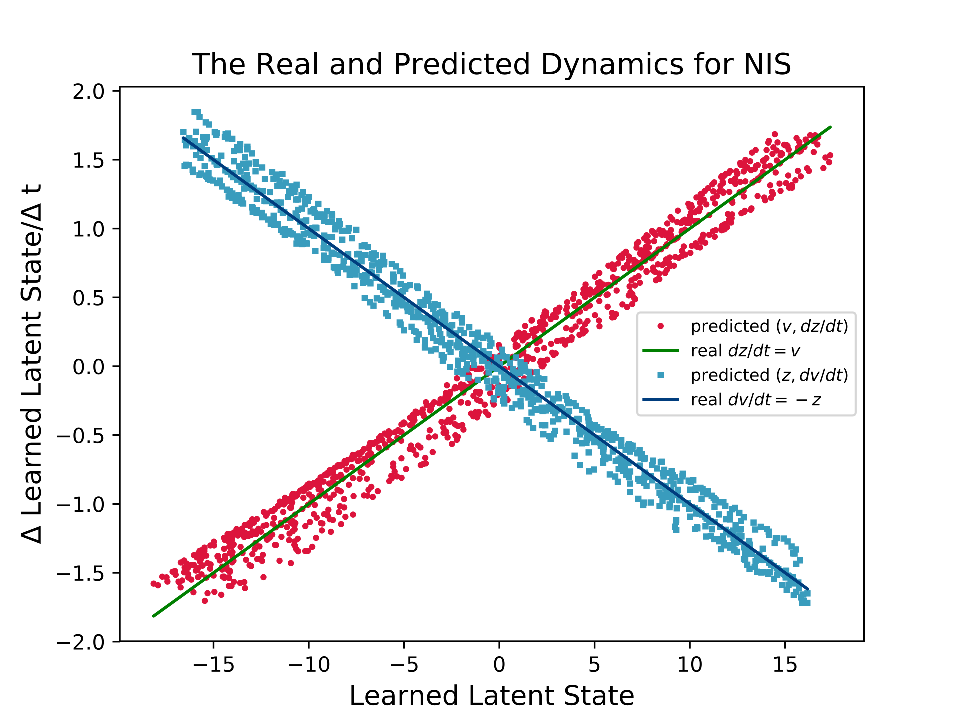}
        \centerline{(c)}
    \end{minipage}
    \begin{minipage}[c]{0.5\textwidth}
        \centering
        \includegraphics[width=1\textwidth]{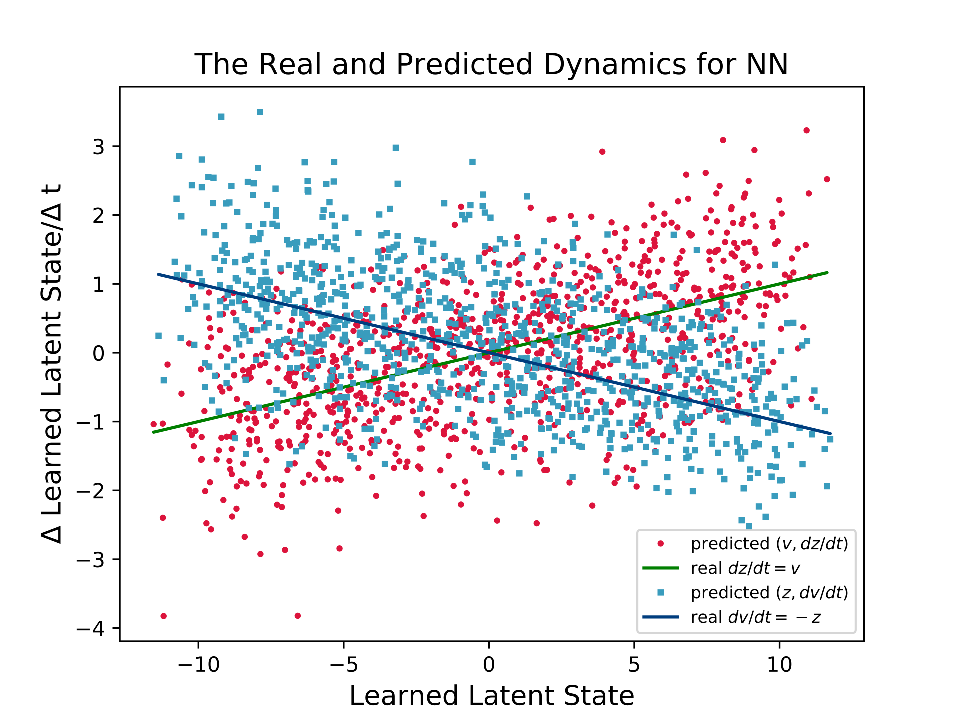}
        \centerline{(d)}
    \end{minipage}
    \begin{minipage}[c]{0.5\textwidth}
        \centering
        \includegraphics[width=1\textwidth]{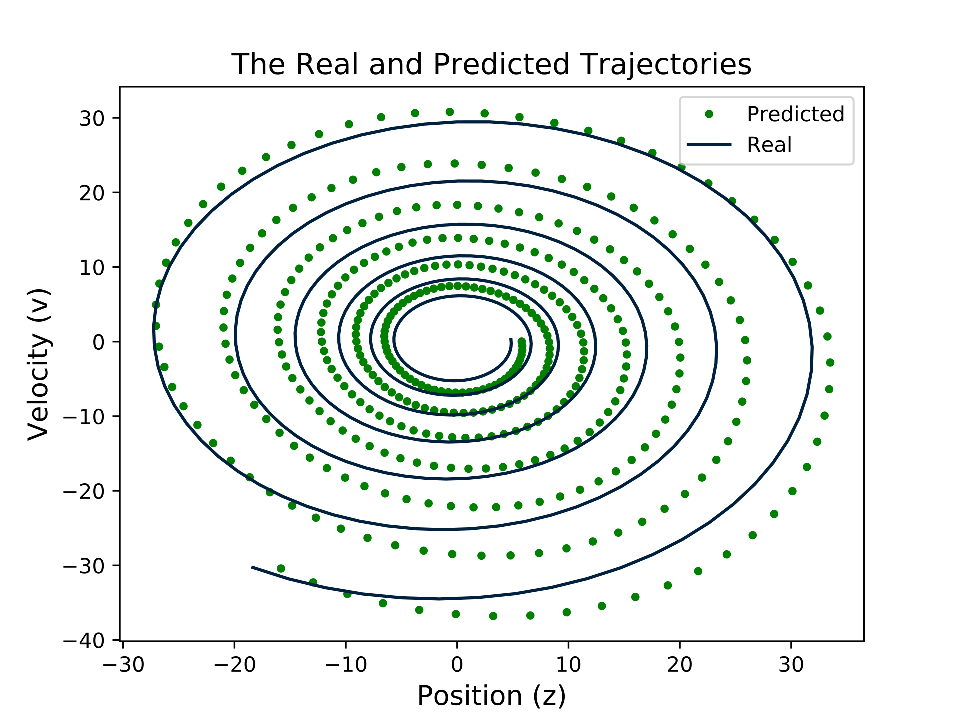}
        \centerline{(c)}
    \end{minipage}
    \begin{minipage}[c]{0.5\textwidth}
        \centering
        \includegraphics[width=1\textwidth]{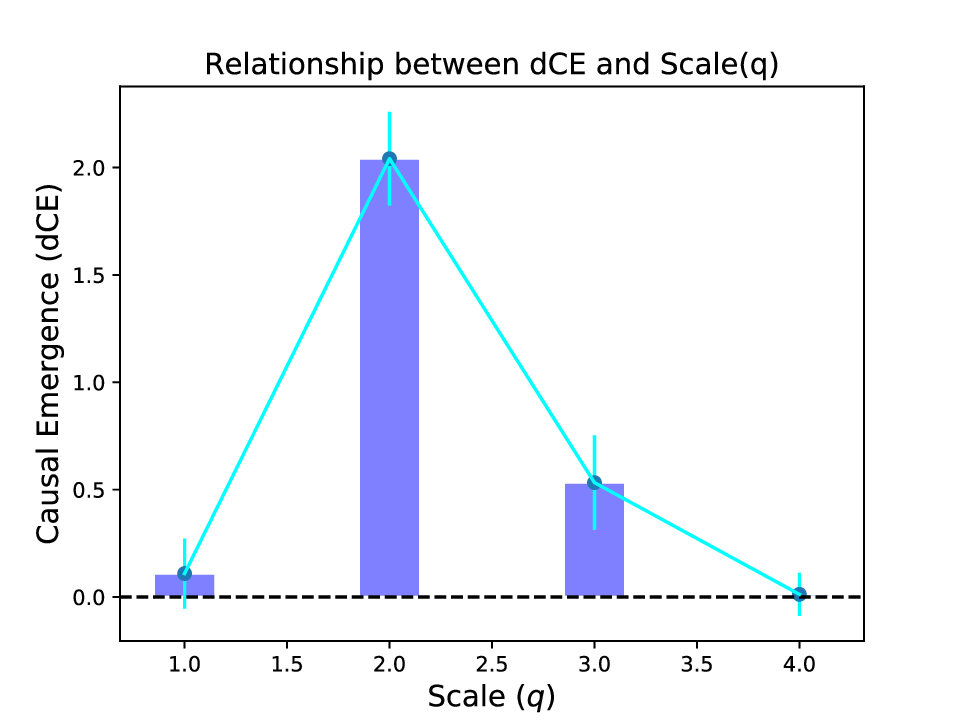}
        \centerline{(d)}
    \end{minipage}
    \caption{Experimentsl Results for the Simple Spring Oscillator with Measurement Noise. We sample data from Equation \ref{eq:spring_oscillator} and \ref{eq:obervation}, and we use Euler method to simulate by taking $dt=0.1$. (a) and (b) show the real macro-state versus the predicted ones both for NIS and the ordinary feed-forward neural network, respectively; (c) and (d) show the real and predicted dynamics, i.e., the dependence between $dz/dt$ and $v$, and $dv/dt$ with $z$ for both neural networks for comparison; (e) shows the real and predicted trajectories with 400 time steps starting from the same latent state; (f) shows the dependence of the dimension averaged of $\textit{Causal Emergence}$ (dCE) on $q$ (the number of effective channels).}
    \label{fig:spring_result}
\end{figure}

The results are shown in Figure \ref{fig:spring_result}. To test if NIS can learn the real latent macro state, we directly plot the predicted and the real latent states. As shown in Figure \ref{fig:spring_result}(a), the predicted and the real curves collapse together which means NIS can recover the macro state in the data although it is unknown. As a comparison, the feed-forward neural network cannot recover the macro state. We can also check if the NIS can learn the dynamic of the macro states by plotting the derivatives of the states ($dz/dt, dv/dt$) against the macro state variables ($v,z$). If the learned dynamics follows Equation \ref{eq:spring_oscillator}, then two cross-over lines for $dz/dt=v$ and $dv/dt=-z$ can be observed as shown in Figure \ref{fig:spring_result}(c). However, the same pattern can not be reproduced on the common feed-forward network as shown in Figure \ref{fig:spring_result}(d). We also test the well-trained NIS by multiple-step prediction as shown in Figure \ref{fig:spring_result}(e). Although there are larger and larger deviation from the prediction and the real data, the general trends can be captured by NIS model. We further study how the dimension averaged causal emergence $dCE$ changes with the scale $q$ which is measured by the number of effective information channels on the well-trained NIS model as shown in Figure \ref{fig:spring_result}(f). $dCE$ peaks at $q=2$ which is exactly same as in the ground truth.

\begin{figure}[htbp]
    \begin{minipage}[c]{0.5\textwidth}
        \centering
        \includegraphics[width=1\textwidth]{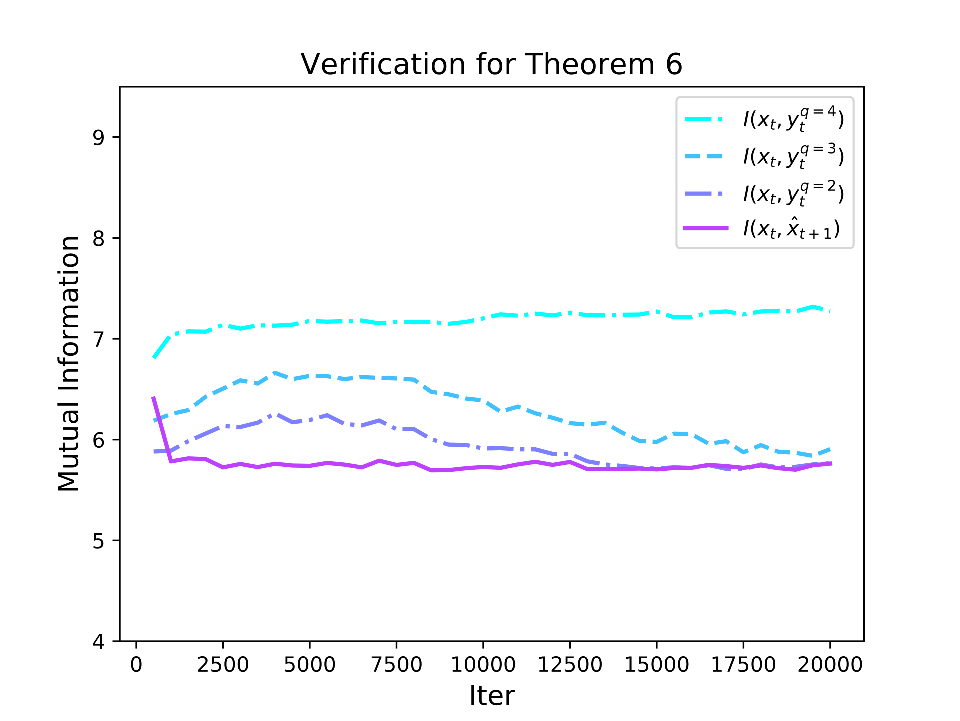}
        \centerline{(a)}
    \end{minipage}
    \begin{minipage}[c]{0.5\textwidth}
        \centering
        \includegraphics[width=1\textwidth]{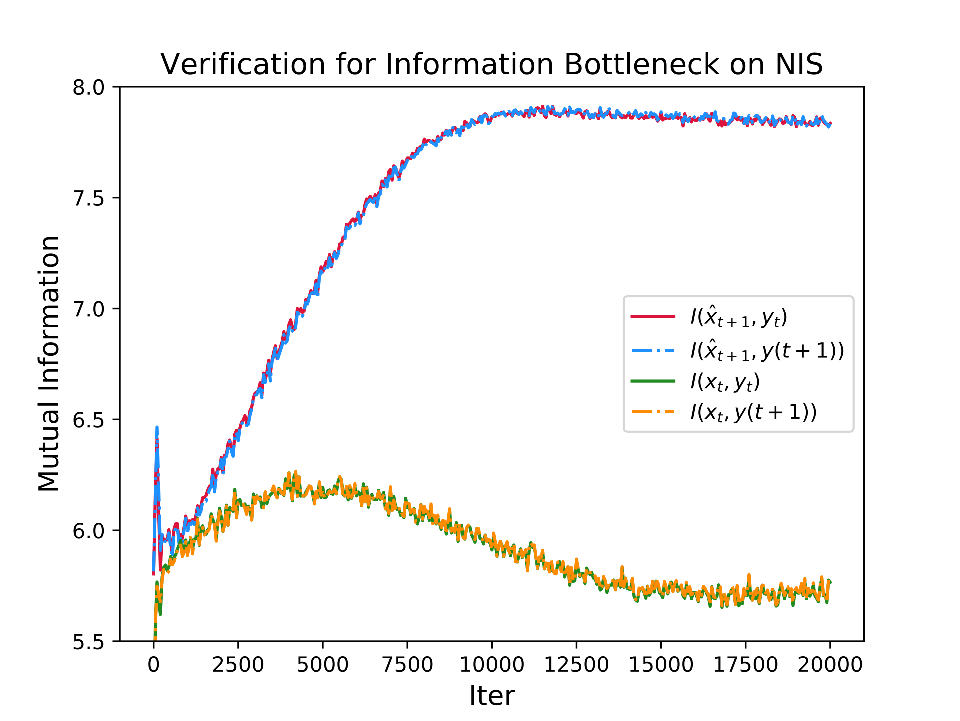}
        \centerline{(b)}
    \end{minipage}
    \begin{minipage}[c]{0.5\textwidth}
        \centering
        \includegraphics[width=1\textwidth]{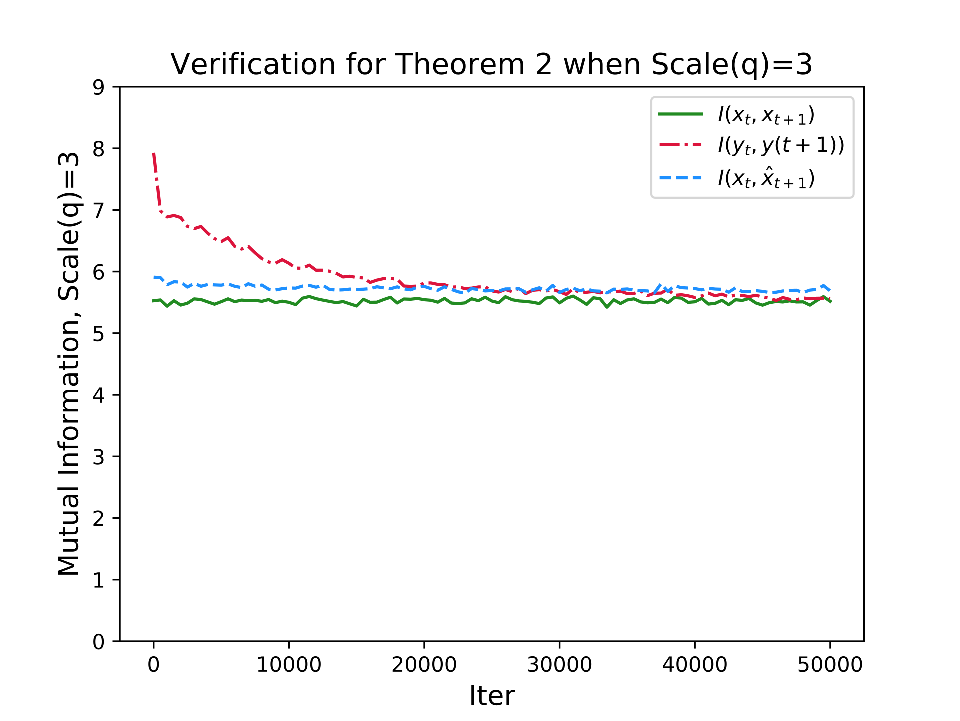}
        \centerline{(c)}
    \end{minipage}
    \begin{minipage}[c]{0.5\textwidth}
    \centering
    \includegraphics[width=1\textwidth]{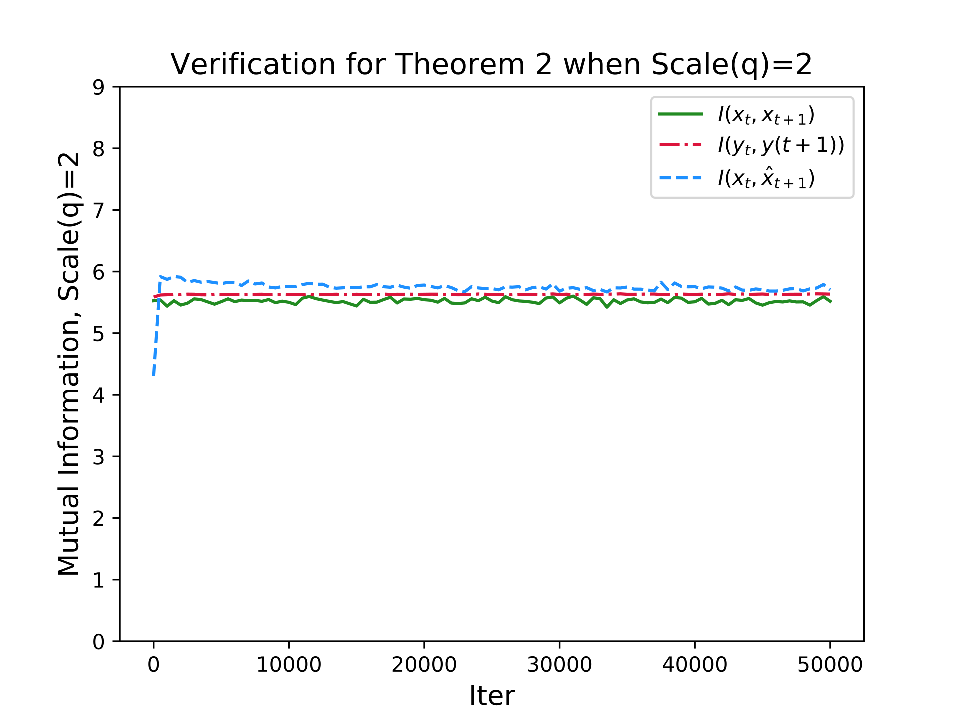}
    \centerline{(d)}
    \end{minipage}
    \caption{Various mutual information between variables change with training iterations. (a) shows the change of mutual information $I(x_t,y_t^{q=4})$, $I(x_t,y_t^{q=3})$, $I(x_t,y_t^{q=2})$ and $I(x_t,\hat{x}_t)$ with the increase of the number of iterations. From the figure, we can see that within the specified number of iterations, $I(x_t,\hat{x}_t)\leq I(x_t,y_t^{q=2})\leq I(x_t,y_t^{q=3})\leq I(x_t,y_t^{q=4})$. Among them, $q$ is the dimension of the coarse-grained system. (b) verifies the theory of information bottleneck on NIS when Scale(q)=2. (c)-(d) show the change of mutual information $I(x_t,x_{t+1})$, $I(y_t,y_{t+1})$ and $I(x_t,\hat{x}_t)$ with the increase of the number of iterations. It can be seen that under different scales, the three mutual information values are close to each other. The standard deviations of the three variables gradually decrease with iteration. So $I(x_t,x_{t+1})\approx I(y_t,y_{t+1})=I(x_t,\hat{x}_t)$  is reflected.}
    \label{fig:spring_Mutual_Information}
\end{figure}

Further, we use experimental results to verify the theorems mentioned in the previous section and the theory of information bottleneck\cite{shwartz2017opening}. First, we show how the mutual information of $I(\mathbf{x}_t,\mathbf{x}_{t+1})$, $I(\mathbf{y}_t,\mathbf{y}(t+1))$, and $I(\mathbf{x}_t,\mathbf{\hat{x}}_{t+1})$ change with time(epoch) when $q$ takes different values as shown in Figures \ref{fig:spring_Mutual_Information}(c) and (d). The results show that all the mutual information converge as predicted by Theorems \ref{thm.info_bottle} and \ref{thm.training}. We also plot the mutual information between $\mathbf{x}_t$ and $\mathbf{y}_t$ with different $q$ to test Theorem \ref{thm.harder}, and the results show that the mutual information increases when $q$ increases as shown in Figure \ref{fig:spring_Mutual_Information}(a).

According to the information bottleneck theory\cite{shwartz2017opening}, the mutual information between latent variable and output may increase while the information between input and latent variable should increase in the early stage and then decrease as training process proceed. As shown in Figure \ref{fig:spring_Mutual_Information}(b), this conclusion is confirmed by the NIS model where the macro-states $\mathbf{y_t}$ and the prediction $\mathbf{y(t+1)}$ are all latent variables. Although the same conclusion is obtained, the information bottleneck can be reflected by the architecture in NIS model much clearer than the general neural networks because $\mathbf{y}_t$ and $\mathbf{y}(t+1)$ is the bottleneck and all other irrelevant information is discarded by the variable $\mathbf{x}_t''$ as shown in Figure \ref{fig:squeezed_channel1}.

\subsubsection{Simple Markov Chain}
\begin{figure}[htbp]
\begin{minipage}[c]{0.5\textwidth}
    \centering
    \includegraphics[width=1\textwidth]{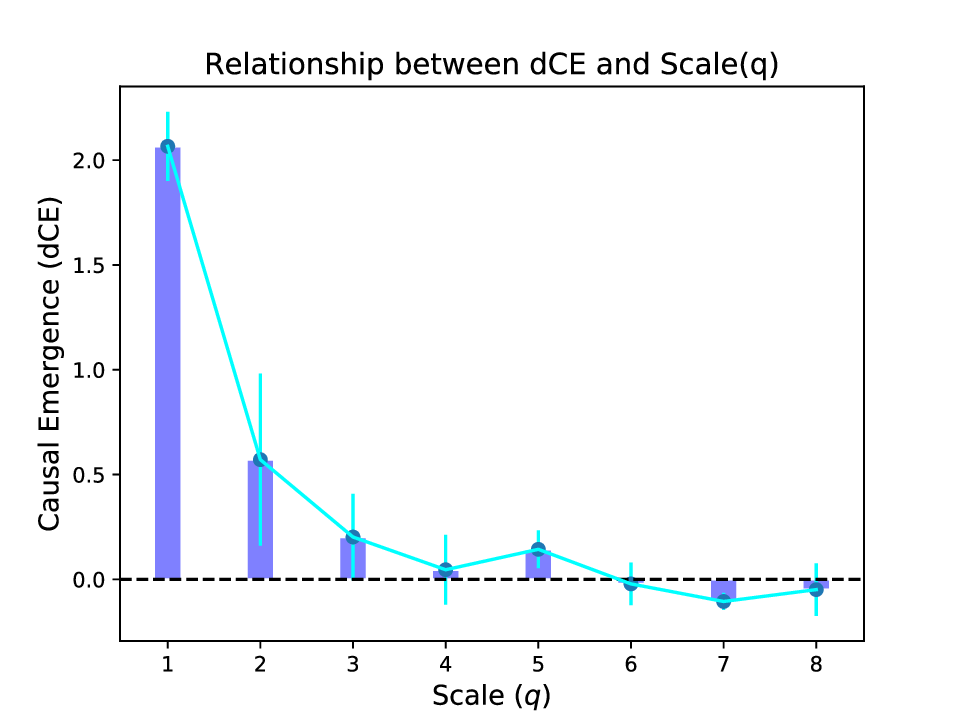}
    \centerline{(a)}
\end{minipage}
\begin{minipage}[c]{0.5\textwidth}
    \centering
    \includegraphics[width=1\textwidth]{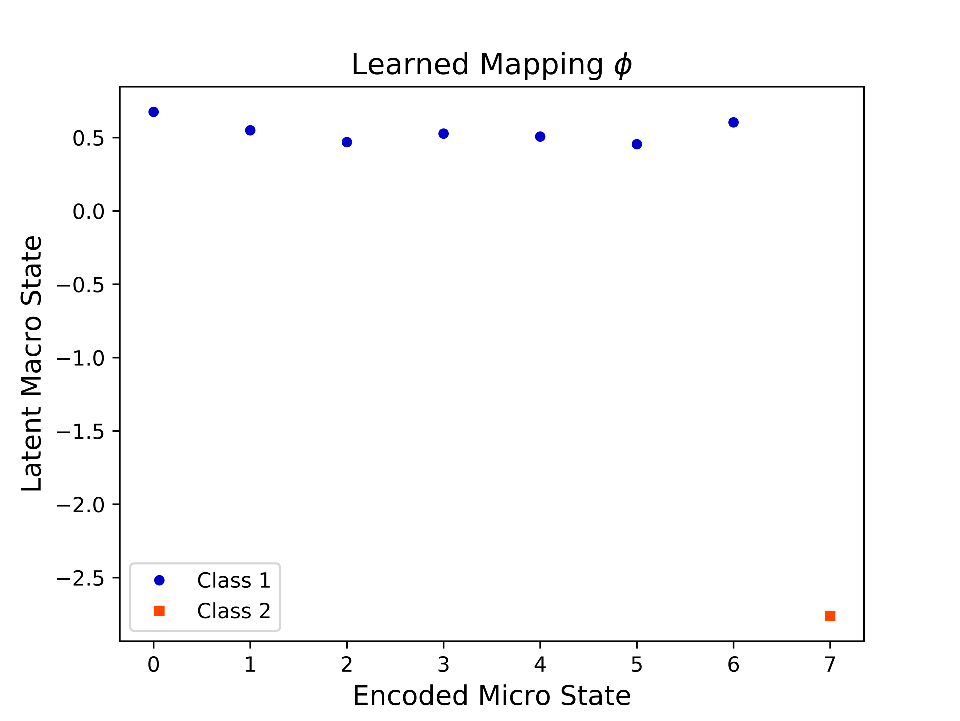}
    \centerline{(b)}
\end{minipage}
\begin{minipage}[c]{0.5\textwidth}
    \centering
    \includegraphics[width=1\textwidth]{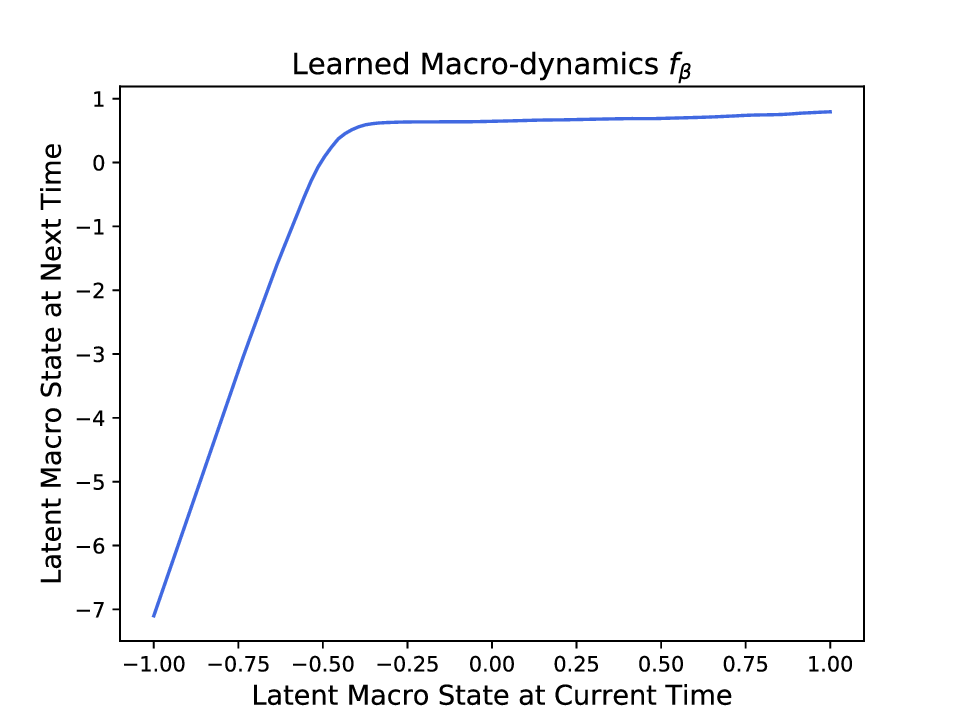}
    \centerline{(c)}
\end{minipage}
\begin{minipage}[c]{0.5\textwidth}
    \centering
    \includegraphics[width=1\textwidth]{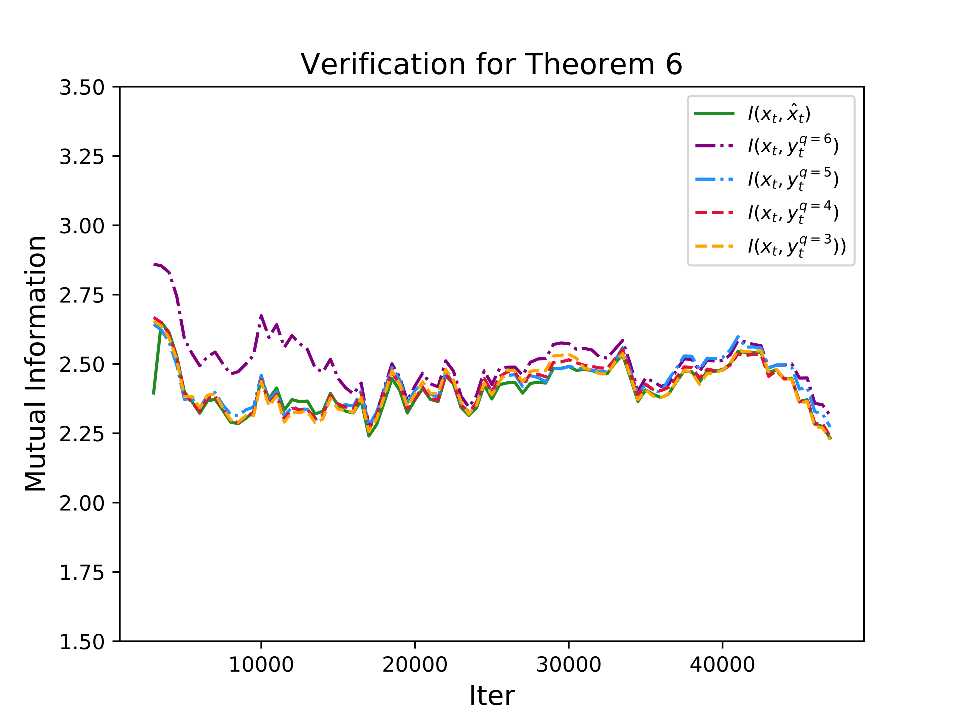}
    \centerline{(d)}
\end{minipage}
    \caption{The dependence of the dimension averaged $\textit{Causal Emergence}$ (dCE) on different scales($q$) of the markov dynamics (a), the learned mapping between micro states and macro states on the optimal scale($q$) (b), and the learned macro-dynamics (the mapping from $\mathbf{y}_t$ to $\mathbf{y}(t+1)$ (c). There are two clear separated clusters on the y-axis in (b) which means the macro states are discrete. We found that the two discrete macro states and the mapping between micro and discrete macro states are identical as the example in ref \cite{Hoel2019map} which means the correct coarse-graining strategy can be discovered by our algorithm automatically under the condition without any prior information. In (d), $I(x_t,\hat{x}_t)\leq I(x_t,y_t^{q=3})\leq I(x_t,y_t^{q=4})\leq I(x_t,y_t^{q=5})\leq I(x_t,y_t^{q=6})$  is reflected. In order to make the data clearer, we have taken a moving average for each group of data. This result can be regarded as the verification of theorem \ref{thm.harder}}
\label{fig:bigmarkov}
\end{figure}

In the second example, we show NIS can work on discrete markov chain, and the coarse-graining strategy can work on state space. The markov chain to generate the data is the following probability transition matrix:
\begin{equation}
    \label{eq:large_markov}
    \begin{pmatrix}
    1/7&1/7&1/7&1/7&1/7&1/7&1/7&0\\
    1/7&1/7&1/7&1/7&1/7&1/7&1/7&0\\
    1/7&1/7&1/7&1/7&1/7&1/7&1/7&0\\
    1/7&1/7&1/7&1/7&1/7&1/7&1/7&0\\
    1/7&1/7&1/7&1/7&1/7&1/7&1/7&0\\
    1/7&1/7&1/7&1/7&1/7&1/7&1/7&0\\
    1/7&1/7&1/7&1/7&1/7&1/7&1/7&0\\
    0&0&0&0&0&0&0&1
    \end{pmatrix}
\end{equation}
The system has 8 states, and seven of them can transfer each other. The last state is standalone. We use a one-hot vector to encode the states. Therefore, for example, state $2$ will be represented as $(0,1,0,0,0,0,0,0)$. We sample the initial state for 50,000 batches to generate data.
We then feed these one-hot vectors into the NIS framework, after training for 50,000 epochs, we can obtain an effective model. The results are shown in Figure \ref{fig:bigmarkov}.

By systematically search for different $q$, we found that the dimension averaged causal emergence(dCE) peaks at $q=1$ as shown in Figure \ref{fig:bigmarkov}(a). On the optimal scale, we can visualize the coarse-graining strategy by Figure \ref{fig:bigmarkov}(b), on which the x-coordinate is the decimal coding for different states, and the y-coordinate represents the coding for the macro-states. We find that the coarse-graining mapping successfully classifies the first seven states into a one macro-state, and leaves the last state stay alone. This learned coarse-graining strategy is identical as the example shown in \cite{Hoel2019map}. 

We also visualize the learned macro-dynamics as shown in Figure \ref{fig:bigmarkov}(c). This is a linear mapping when $y_t<0$ and almost a constant for $y_t>0$. Therefore, the dynamics can guarantee that all the first seven micro-states can be separated with the last state. We also verify Theorem \ref{thm.info_bottle} in Figure \ref{fig:bigmarkov}(d).
\subsubsection{Simple Boolean Network}
Our framework can not only work on continuous time series and markov chain, but also can work on a networked system on which each node follows a discrete micro mechanism.

\begin{figure}
    \centering
    \includegraphics[scale=0.7]{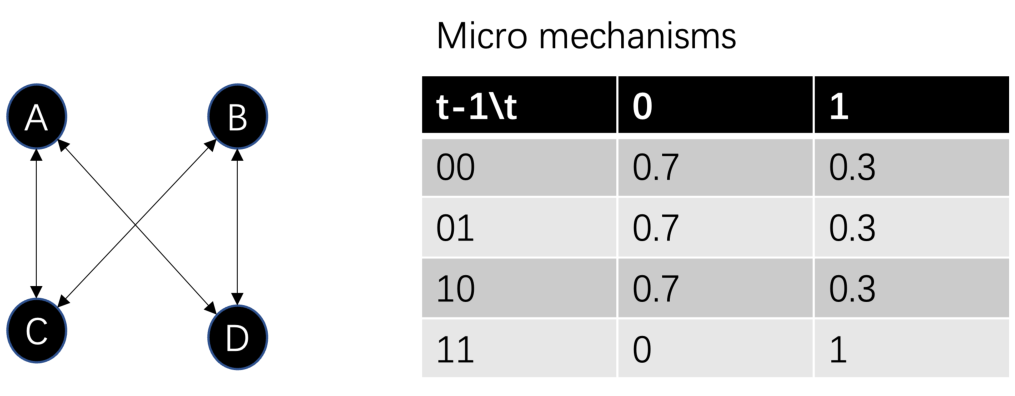}
    \caption{An exampled Boolean network(left) and its micro mechanisms on nodes(right). Each node's state on the next time step is affected by its neighboring nodes' state combination randomly. The transition probabilities (micro mechanisms) on each case are shown in the table.}
\label{fig:micro_mechanism}
\end{figure}

For example, boolean network is a typical discrete dynamical system in which the node contains two possible states (0 or 1), and the state of each node is affected by the state of the neighbors connected to it. We follow the example in \cite{Hoel2013causal_emergence}. Figure \ref{fig:micro_mechanism} shows an exampled boolean network with 4 nodes, and each node follows the same micro mechanism as shown in the table of Figure \ref{fig:micro_mechanism}. In the table, each entry is the probability of each node's state conditions on the state combination of its neighbors. For example, if the current node is A, then the first entry is  $Pr(x_A^{t+1}=0|x_C^t=0, x_D^t=0)=0.7$, which means that A will take value 0 with probability 0.7 when the state combination of C and D is 00. By taking all the single node mechanisms together, we can obtain a large markovian transition matrix with $2^4=16$ states which is the complete micro mechanism of the whole network.

\begin{figure}[htbp]
    \begin{minipage}[c]{0.5\textwidth}
        \centering
        \includegraphics[width=1\textwidth]{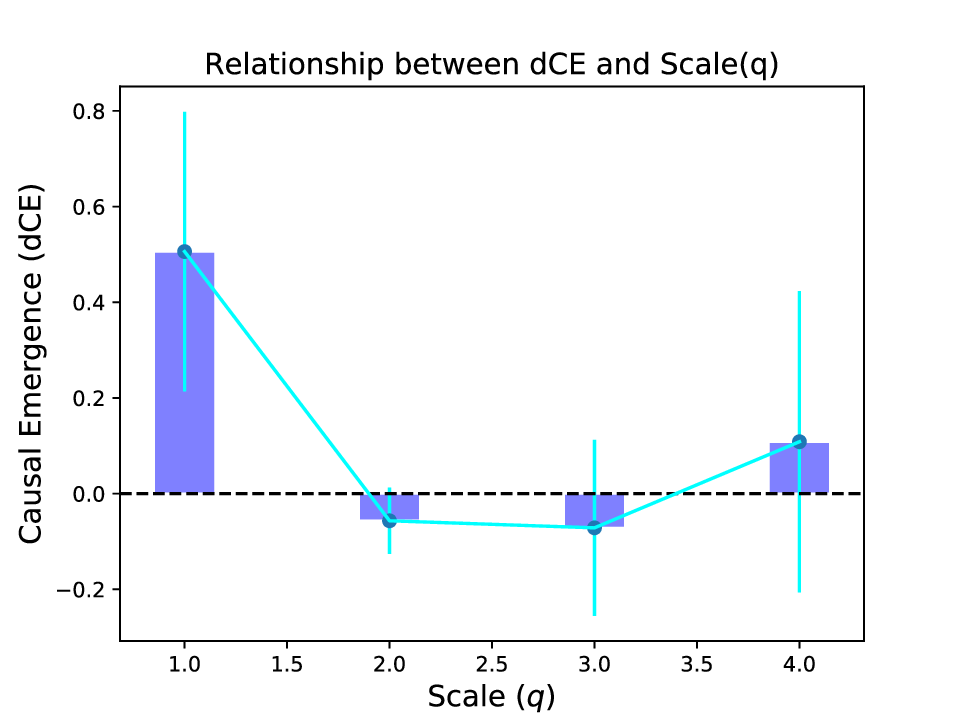}
        \centerline{(a)}
    \end{minipage}
     \begin{minipage}[c]{0.5\textwidth}
        \centering
        \includegraphics[width=1\textwidth]{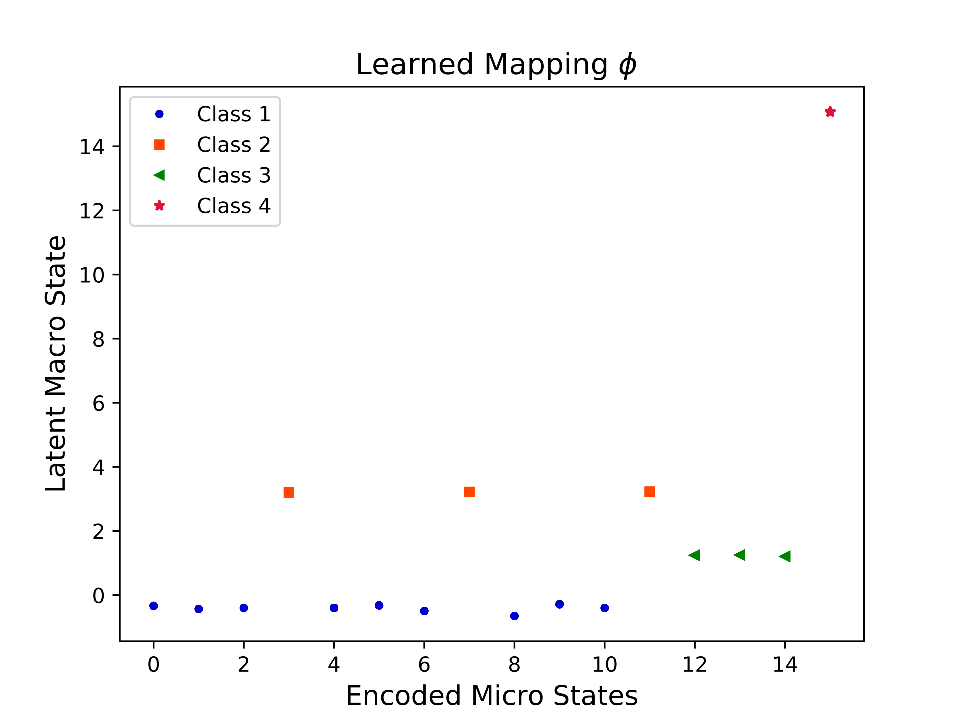}
        \centerline{(b)}
    \end{minipage}
    \begin{minipage}[c]{0.5\textwidth}
        \centering
        \includegraphics[width=1\textwidth]{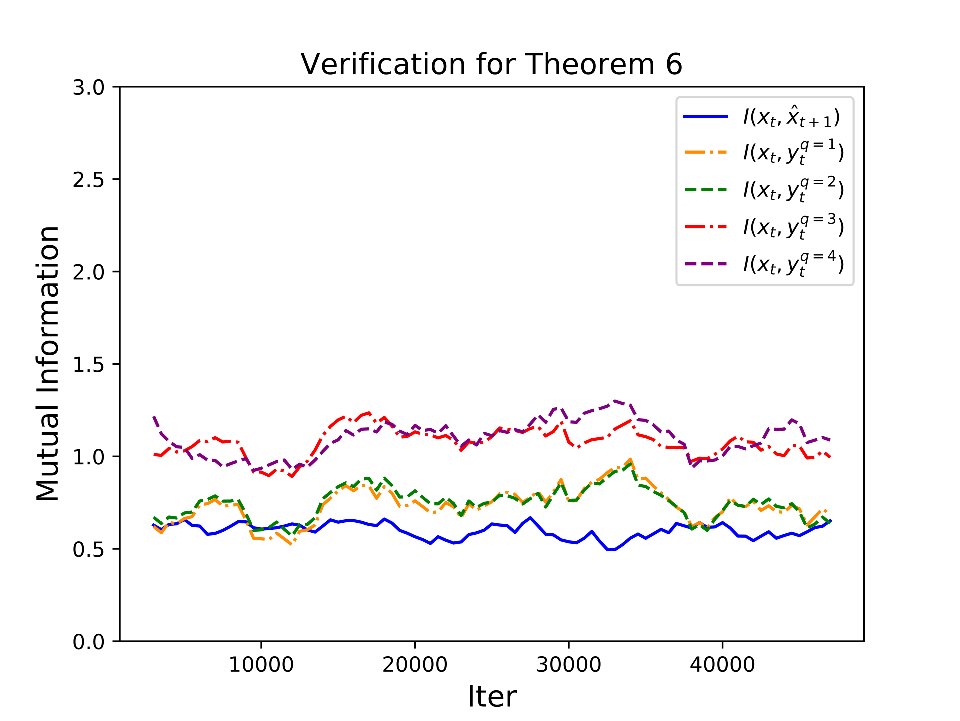}
        \centerline{(c)}
    \end{minipage}
    \begin{minipage}[c]{0.5\textwidth}
        \centering
        \includegraphics[width=1\textwidth]{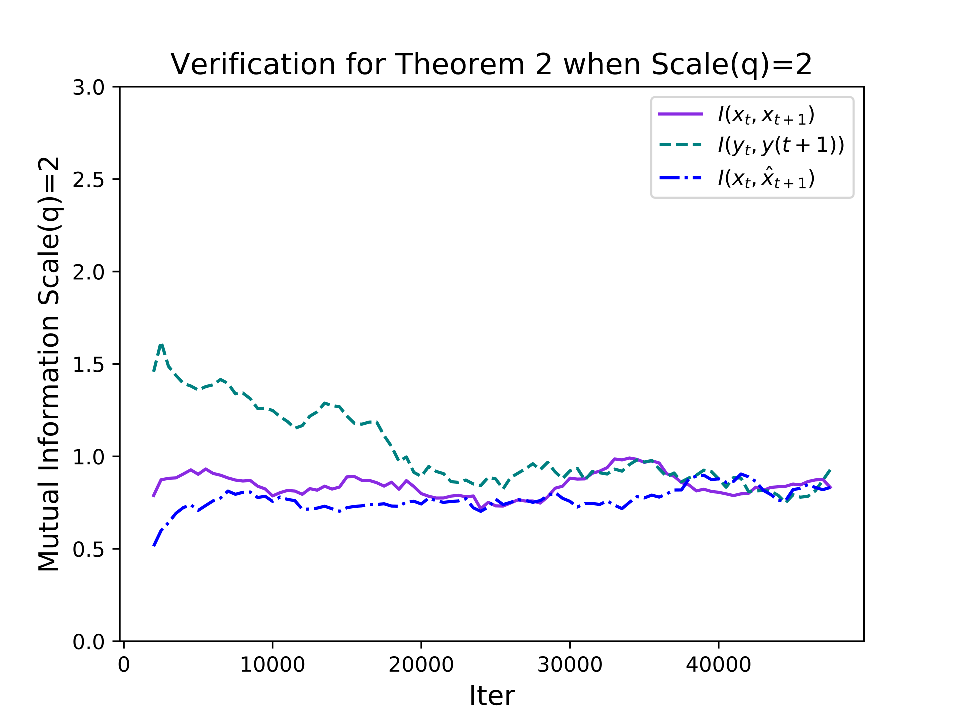}
        \centerline{(d)}
    \end{minipage}
    \caption{Experimentsl Results for the Boolean Network. The dependence of the dimension averaged $\textit{Causal Emergence}$ (dCE) on different scales($q$) (a) and the learned mapping between micro states and macro states on the optimal scale($q$) (b). There are four clear separated clusters on the y-axis in (b) which means the macro states are discrete. We found that the four discrete macro states and the mapping between micro and discrete macro states are identical as the example in ref \cite{Hoel2013causal_emergence} which means the correct coarse-graining strategy can be discovered by our algorithm automatically under the condition without any prior information. (c) shows the change of mutual information $I(x_t,y_t^{q=4})$, $I(x_t,y_t^{q=3})$, $I(x_t,y_t^{q=2})$,$I(x_t,y_t^{q=1})$ and $I(x_t,\hat{x}_t)$ with the increase of the number of iterations. From the figure, we can  see approximately that within the specified number of iterations, $I(x_t,\hat{x}_t)\leq I(x_t,y_t^{q=1})\leq I(x_t,y_t^{q=2})\leq I(x_t,y_t^{q=3})\leq I(x_t,y_t^{q=4})$. (d) shows the change of mutual information $I(x_t,x_{t+1})$, $I(y_t,y_{t+1})$ and $I(x_t,\hat{x}_t)$ with the increase of the number of iterations. In order to easily observe the trend of data changes, we added a moving average curve for each group of data. It can be seen that under different scales, the three mutual information values are close to each other. $I(x_t,x_{t+1})\approx I(y_t,y_{t+1})=I(x_t,\hat{x}_t)$ is reflected. Considering the experimental error, the overall trend of the data still conforms to the theorem.}
    \label{fig:boolnet_result}
\end{figure}

We sample the one step state transition of the entire network for 50,000 batches and each batch contains 100 different initial conditions which are randomly sampled from the possible state space evenly, and we then feed these data to the NIS model. By systematically search for different $q$, we found that the dimension averaged causal emergence peaks at $q=1$ as shown in Figure \ref{fig:boolnet_result}(a). Under this condition, we can visualize the coarse-graining strategy by Figure \ref{fig:boolnet_result}(b), on which the x-coordinate is the decimal coding for the binary micro-states (e.g., 5 denotes for the state 0101), and the y-coordinate represents the codes for macro-states. The data points can be clearly classified into 4 clusters according to their y-coordinate. This means the NIS network found 4 discrete macro-states although the states are continuous real numbers. Interestingly, we found that the mapping between the 16 micro states and 4 macro states are identical as the coarse-graining strategy shown in the example in ref \cite{Hoel2013causal_emergence}. However, any prior information neither the method on how to group the nodes nor the coarse graining strategy, nor the dynamics are known by our algorithm. Finally, theorems \ref{thm.info_bottle} and \ref{thm.harder} are verified in this example as shown in Figure \ref{fig:boolnet_result} (c) and (d).


\section{Concluding Remarks}
In this paper, we propose a novel neural network framework, Neural Information Squeezer, for discovering coarse-graining strategy, macro-dynamic and emergent causality in time series data. We first define effective coarse-graining strategy and macro-dynamic by constraining the coarse-graining strategies to predict the future micro-state with a precision threshold. And then, the causal emergence identification problem can be understood as a maximization problem for effective information under the constraint. 

We then use an invertible neural network incorporating with the projection operation to realize the coarse-graining strategy. The usage of invertible neural network can not only allow us to reduce the number of parameters by sharing them between the encoder and the decoder but also can facilitate us to analyze the mathematical properties of the whole NIS architecture. 

By treating the framework as a squeezed information channel, we can prove four important theorems. The results show that if the causal connection in the data is strong, then as we train the neural networks, the macro-dynamics will increase its informativeness. And during this process, the determinant of the Jacobian of the bijector will increase in the same time. We also found a mathematical expression for the effective information of the macro-dynamics without the explicit dependence on the macro-dynamics, and it is determined solely by the bijector and the data when the whole framework is well trained. Furthermore, if the framework has been trained in a sufficient time, the mutual information of the macro-dynamics will keep a constant no matter the scale $q$ is. However, as $q$ decreases, the mutual information or the bandwidth on the encoder part also decreases and closed to the information limitation on the entire channel such that it can make correct prediction for the future micro-states. Thus, the task becomes harder for the encoder because more effective information must be encoded and pass to the dynamics learner such that it can make correct prediction with less information. Numerical experiments show that our framework can reconstruct the dynamics in different scales and also can discover emergent causality in data on several classic causal emergence examples.

There are several weak points in our framework. First, it can only work on small data set. The major reason is the invertible neural network is very difficult to train on large data set. Therefore, we will use some special techniques to optimize the architecture in future. Second, the framework is still lack of explainability, the grouping method for variables is implicitly encoded in the invertible neural network although we can illustrate what the coarse-graining mapping is, and decompose it into information conversion and information discarding parts clearly. A more transparent neural network framework with more explanatory power is deserved for future studies. Third, the conditional distribution that the model can predict actually is limited as Gaussian or Laplacian, and it should be extended to more general distributional forms in future studies.

There are several theoretical problems left for future studies. For example, we conjecture that all coarse-graining strategies can be decompose into a bijection and a projection, but this needs strict mathematical proof. Second, although an explicit expression for EI on macro-dynamics has been derived under NIS, we still cannot directly predict the causal emergence in the data. We believe that a more concise analytic results on the EI should be derived by setting some constraints on the data. Furthermore, we think the meaning and the usage of the discarding variable $\mathbf{x}_t''$ should be further explored since that it may relate with the redundant information of a pair of variables toward a target\cite{Williams2017PID}. Therefore, we guess more deep connections between the framework of NIS and the mutual information decomposition may exist and NIS may work as a numeric tool to decompose the mutual information.

\vspace{6pt} 



\authorcontributions{Conceptualization and methodology, J.Z.; coding, J.Z. and K.L.; writing, J.Z. and K.L. All authors have read and agreed to the published version of the manuscript.}

\funding{This research was funded by by the National Natural Science Foundation of China (NSFC) under Grant No. 61673070 at \url{https://www.nsfc.gov.cn/}}

\dataavailability{All the codes and data are available at: \url{https://github.com/jakezj/NIS_for_Causal_Emergence}} 

\acknowledgments{J.Z. acknowledge the discussion in the reading group on ``Causal Emergence'' organized by Swarma Club, particularly for Prof. Everrete You and Erik Hoel.}

\appendixtitles{no} 
\appendixstart
\appendix
\section[\appendixname~\thesection]{RealNVP Implementation of Invertible Neural Network}
\label{sec.realnvp}
In the main text, we mentioned that the invertible neural network can be realized by a RealNVP module. The architecture of a RealNVP module can be visualized by Figure \ref{fig:realvnp}. Concretely, if the input of the module is $\mathbf{x}$ with dimension $n$ and the output is $\mathbf{x'}$ with the same dimension, then the RealNVP module can perform the following computation steps:

\begin{equation}
    \label{eq.x1x2}
    \left\{
    \begin{aligned}
        &\mathbf{x_1} = \mathbf{x}_{1:m}\\
        &\mathbf{x_2} = \mathbf{x}_{m:n}
    \end{aligned}
    \right.
\end{equation}
where, $m$ is an integer in between 1 and $n$.
\begin{equation}
    \label{eq:sandt}
    \left\{
    \begin{aligned}
        &\mathbf{x'_1} = \mathbf{x_1}\bigotimes s_1(\mathbf{x_2})+t_1(\mathbf{x_2})\\
        &\mathbf{x'_2} = \mathbf{x_2}\bigotimes s_2(\mathbf{x'_1})+t_2(\mathbf{x'_1})
    \end{aligned}
    \right.
\end{equation}
where, $s_1,s_2,t_1$ and $t_2$ are feed-forward neural networks with arbitrary architectures, while their input-output dimensions must match with the data. In practice, $s_1$ or $s_2$ always do an exponential operation on the output of the feed-forward neural network \cite{dinh2016density} to facilitate the inverse computation. 

Finally,
\begin{equation}
    \mathbf{x'}=\mathbf{x'_1}\bigoplus \mathbf{x'_2}
\end{equation}
It is not difficult to verify that all three steps are invertible. Equation \ref{eq:sandt} is invertible because the same form but with negative signs can be obtained by solving the expressions of $\mathbf{x_1}$ and $\mathbf{x_2}$ with $\mathbf{x'_1}$ and $\mathbf{x'_2}$ from Equation \ref{eq:sandt}.

To simulate more complex invertible functions, we always duplex the basic RealNVP modules by stacking them together. In the main text, we use duplex the basic RealNVP module by three times.
\section[\appendixname~\thesection]{Approximated Calculation of Effective Information for Neural Networks}
\label{sec.gaussianei}

In this paper, we propose an approximated method to calculate EI for a neural network. Conventional methods usually coarse-grain the input and output spaces into small regions, and estimate the probability of each region by the frequency. However, this estimation is inaccurate especially for the regions with small probability. 

To avoid this problem, we propose a new method to estimate the mutual information of a neural network. The key idea is to treat a Neural Network as a conditional probability with a Gaussian(or Laplacian) distribution in which the mean value is the output vector of the neural network, and the standard deviation takes the Mean Square Error of the prediction. The concrete distributional form (Gaussian or Laplacian) is determined by the types of Loss function. If MSE(Mean Square Error) is taken then Gaussian distribution is considered, otherwise if MAE(Mean Absolute Error) is considered then Laplacian distribution is considered. Without lose genarlity, here, we take the distributional form as Gaussian.

We restate theorem \ref{thm.ei_gauss}:

\textbf{Theorem \ref{thm.ei_gauss}} In general, if the input of a neural network is $X=(x_1,x_2,\cdot\cdot\cdot,x_n)\in [-L,L]^n$, where $L$ is a big integer, the output is $Y=(y_1,y_2,\cdot\cdot\cdot,y_m)$, and $Y=\mu(X)$. Here $\mu$ is the deterministic mapping implemented by the neural network: $\mu: \mathcal{R}^n\rightarrow \mathcal{R}^m$, and its Jacobian matrix at $X$ is $\partial_{X'} \mu(X)\equiv \left\{\frac{\partial \mu_i(X')}{\partial X'_j}\left|_{X'=X}\right.\right\}_{nm}$, and if the neural network can be regarded as an Gaussian distribution conditional on given $X$:
\begin{equation}
    \label{eq:gaussian}
    p(Y|X)=\frac{1}{\sqrt{(2\pi)^m|\Sigma|}}\exp{\left(-\frac{1}{2}(Y-\mu(X))^T\Sigma^{-1}(Y-\mu(X))\right)}
\end{equation}
where, $\Sigma=diag(\sigma_1^2,\sigma_2^2,\cdot\cdot\cdot,\sigma_m^2)$ is the co-variance matrix, and $\sigma_i$ is the standard deviation of the output $y_i$ which can be estimated by the mean square error of $y_i$. Then the effective information (EI) of the neural network can be calculated in the following way:

(i) If there exists $X$ such that $\det(\partial_{X'} \mu(X))\neq 0$, then the effective mutual information (EI) can be calculated as:
\begin{equation}
\label{eq.ei_appendix}
\begin{aligned}
        EI_L(\mu)=I(do(X\sim U([-L,L]^{n};Y)\approx & -\frac{m+m \ln (2\pi)+ \sum_{i=1}^{m}\ln \det(\sigma_i^2)}{2}\\
        & + n \ln (2L) + \mathbb{E}_{X\sim U([-L,L]^n} \left(\ln |\det(\partial_{X'} \mu(X))|\right).
\end{aligned}
\end{equation}
where, $U([-L,L]^n)$ is the uniform distribution on $[-L,L]^n$, and $|\cdot|$ is absolute value, and $\det$ is determinant. 

(ii) If $\det(\partial_{X'} \mu(X))\equiv 0$ for all $X$, then $EI\approx 0$
\begin{proof}
Because the calculation of mutual information can be separated into two parts:
\begin{equation}
\label{eq:mutualinformation}
\begin{aligned}
        I(X;Y)&=\int_{XY}p(X,Y)\ln\frac{p(X,Y)}{p(X)p(Y)}dXdY=\int_{XY}p(X)p(Y|X)\ln\frac{p(Y|X)}{p(Y)}dYdX\\
        &=\int_X p(X)\left(\int_Y p(Y|X)\ln p(Y|X)dY\right)dX\\
        &-\int_X p(X)\left(\int_Y p(Y|X)\ln p(Y)dY\right)dX
\end{aligned}
\end{equation}

By inserting Equation \ref{eq:gaussian} into Equation \ref{eq:mutualinformation}, the first term becomes(the Shannon entropy of the Gaussian Distribution):
\begin{equation}
    \label{eq:first_term}
    \int_Y p(Y|X)\ln p(Y|X) dY=-\frac{m+m \ln (2\pi)+\ln \det(\Sigma) }{2}
\end{equation}

However, it is hard to derive an explicit expression of the second term in Equation \ref{eq:mutualinformation} because it contains integration. So we can expand $\mu(X')$ into Taylor series on the point $X$ and keep only the first order term:
\begin{equation}
    \mu(X')\approx \mu(X)+\partial_{X'}\mu(X)(X'-X),
\end{equation}
where $\partial_{X'}\mu(X)\equiv \frac{\partial \mu(X')}{\partial X'}\left|_{X'=X}\right.=\left(\frac{\partial \mu_i(X)}{\partial X'_j}\right)_{m,n}$.

(i) If there exists $X$: $|\det(\partial_{X'}\mu(X))|\neq 0$, thus:
\begin{equation}
\begin{aligned}
    p(Y)&=\int_{X'} p(X',Y)dX'=\int_{X'} p(X')p(Y|X')dX'=\int_{X'} \rho p(Y|X')dX'\\&\approx \rho \frac{1}{\sqrt{|\Sigma|}}\frac{1}{\sqrt{|\det\left(\partial_{X'}\mu(X)^T\cdot\Sigma^{-1}\cdot\partial_{X'}\mu(X)\right)|}}=\frac{\rho}{|\det\left(\partial_{X'}\mu(X)\right)|},
\end{aligned}
\end{equation}
where, $\rho=(2L)^{-n}$.
This is the multivariate Gaussian integral. Therefore:
\begin{equation}
    \label{eq:second_term}
    \begin{aligned}
           \int_X\left(\int_Y p(Y|X)\ln p(Y)dY\right)dX&\approx \ln \rho -\rho\int_X \ln |\det\left(\partial_{X'}\mu(X)\right)| dX\\
           &\approx \ln \rho -\mathbb{E}_{X\sim U([-L,L]^n)} \ln |\det\left(\partial_{X'}\mu(X)\right)| 
    \end{aligned}
\end{equation}
Thus, EI can be derived by combining the two terms together:
\begin{equation}
\label{eq.eicompute}
    EI_L\approx -\frac{m+m \ln (2\pi) + \ln |\det\left(\Sigma\right)|}{2}-\ln \rho +\mathbb{E}_{X\sim U([-L,L]^n)} \ln |\det\left(\partial_{X'}\mu(X)\right)|
\end{equation}

To insert $\rho=(2L)^{-n}$ and $\det(\Sigma)=\prod_{i=1}^m\sigma_i$ into Equation \ref{eq.eicompute}, we obtain Equation \ref{eq.ei_appendix}.
(ii) If $\det\left(\partial_{X'}\mu(X)\right)=0$ for all $X$, which means $\mu(X)\equiv \mu_0$ where $\mu_0$ is a constant, then:
\begin{equation}
    p(Y)\approx \frac{\exp\left(-\frac{1}{2}(Y-\mu_0)^T\Sigma^{-1}(Y-\mu_0)\right)}{\sqrt{(2\pi)^m|\det(\Sigma)|}},
\end{equation}
so,
\begin{equation}
    \int_Y p(Y|X)\ln p(Y)dY\approx -\frac{m+m\ln \left(2\pi\right) + \ln{|\det(\Sigma)|}}{2},
\end{equation}
Combining with Equation \ref{eq:first_term}, we have:
\begin{equation}
    EI_L\approx 0
\end{equation}
\end{proof}

With this theorem, we can numerically calculate the EI of a neural network in an approximate way. The mathematical expectation can be approximated by averaging the logarithm of the determinant of the Jacobian on the samples of $X$ drawn on the hyper-cube $[-L,L]^n$ uniformly. This method can avoid partitioning intervals and counting frequencies which are very difficult when the dimension is large.

\begin{lem}
\label{lemma.projection}
(Projection does not affect mutual information): Suppose $X\in \mathcal{R}^{p+q}$ and $Y\in Dom(Y)$, where $p,q\in \mathcal{Z}^+$. And $X$ can be decomposed as two components $U\in \mathcal{R}^p, V\in \mathcal{R}^q$, that is:
\begin{equation}
    X=U\bigoplus V,
\end{equation}
where, $\bigoplus$ represents vector concatenation. We call that $U$ and $V$ are $X$'s projections on $p$ or $q$ dimensional sub spaces, respectively. If $U$ and $Y$ form a markov chain $U\rightarrow Y$, and $V$ is independent on $Y$, then we have:
\begin{equation}
    I(X;Y)=I(U;Y)
\end{equation}
\end{lem}

\begin{proof}
Notice that the joint distribution of $X$ and $Y$ can be written as:
\begin{equation}
    p(X,Y)\equiv p(U\bigoplus V,Y)\equiv p(U,V,Y)
\end{equation}
Further, because $X\rightarrow U$ forms a Markov chain, but $V$ is not, thus:
\begin{equation}
    p(V,Y|U)=p(V|U)p(Y|U,V)=p(V|U)p(Y|U),
\end{equation}
thus, we have:
\begin{equation}
    p(X,Y)=p(U,V,Y)=p(U)p(V,Y|U)=p(U)p(V|U)p(Y|U)
\end{equation}
Therefore:
\begin{equation}
\label{eq.theorem2}
\begin{aligned}
    I(X;Y)&=\int_{X,Y} p(X,Y)\ln \frac{p(X,Y)}{p(X)p(Y)}dXdY\\
    &=\int_{U,V,Z}p(U)p(V|U)p(Y|U)\ln\frac{p(U)p(V|U)p(Y|U)}{p(U)p(V|U)p(Y)}dUdVdZ\\
    &=\int_{U,Y}p(U)p(Y|U)\ln\frac{p(U)p(Y|U)}{p(U)p(Y)}dUdY\\
    &=I(U;Y)
\end{aligned}
\end{equation}
\end{proof}

\begin{lem}
\label{lemma.concatenation}
(Mutual information will not be affected by concatenating independent variables): If $X\in Dom(X)$ and $Y\in Dom(Y)$ form a markov chain $X\rightarrow Y$, and $Z\in Dom(Z)$ is a random variable which is independent on both $X$ and $Y$, then:
\begin{equation}
    I(X;Y)=I(X;Y\bigoplus Z)
\end{equation}
\end{lem}

\begin{proof}
Because:
\begin{equation}
    p(X,Y\bigoplus Z)=p(X,Y,Z)=p(X)p(Y,Z|X),
\end{equation}
furthermore, because $X\rightarrow Y$ and $Z$ is independent on both $X$ and $Y$, therefore:
\begin{equation}
    p(Y,Z|X)=p(Y|X)p(Z).
\end{equation}
Thus:
\begin{equation}
\begin{aligned}
    I(X;Y\bigoplus Z)&=I(X;Y,Z)=\int_{XYZ}p(X,Y,Z)\log \frac{p(X,Y,Z)}{p(X)p(Y,Z)}dXdYdZ\\
    &=\int_{XYZ}p(X)p(Y,Z|X)\ln \frac{p(X)p(Y,Z|X)}{p(X)p(Y)p(Z)}dXdYdZ\\
    &=\int_{XYZ}p(X)p(Y|X)p(Z)\ln \frac{p(X)p(Y|X)}{p(X)p(Y)}dXdYdZ\\
    &=\int_{XY}p(X,Y)\ln \frac{p(X,Y)}{p(X)p(Y)}dXdY=I(X;Y)\\
\end{aligned}
\end{equation}
\end{proof}

\begin{figure}
    \centering
    \includegraphics[scale=0.5]{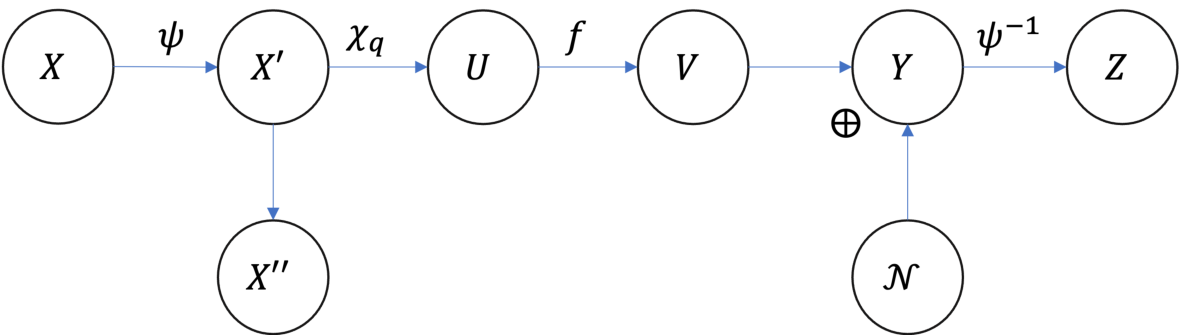}
    \caption{The graphic model of a squeezed information channel}
    \label{fig:squeezed_channel}
\end{figure}

\begin{dfn}
(Squeezed Information Channel):
A squeezed information channel is a graphic model as shown in Figure \ref{fig:squeezed_channel} which also satisfies the following requirements: (1) the mapping $\psi$ from $X$ to $X'$ is a bijection; (2) $\chi_q$ is a $q$ dimensional projector, that is $U$ is a $q$ dimensional projection of $X'$; (3) $U$ and $V$ form a Markov chain $U\rightarrow V$, and $f$ is the conditional probability $P(V|U)$; (3) $\mathcal{N}$ is a random noise which is independent on all other variables; (4) $Y=V\bigoplus N$.
\end{dfn}

It is not hard to know that Figure \ref{fig:architecture} is actually a special case of the squeezed information channel, where $\mathbf{y}_t$ , $\mathbf{y}(t+1)$, and $\hat{\mathbf{x}}_{t+1}$ correspond to $U$, $V$, and $Z$ respectively. 

For this general graphic model, we can prove theorem \ref{thm.info_bottle}:

\textbf{Theorem \ref{thm.info_bottle}(Information bottleneck of the Squeezed Information Channel)}: For the squeezed information channel as shown in Figure \ref{fig:squeezed_channel} and for any $\psi, f$ and $\mathcal{N}$, we have: 
\begin{equation}
    I(X;Z)=I(U;V)
\end{equation}
\begin{proof}
Because $\psi$ and $\psi^{-1}$ are all one to one mappings, thus, according to Lemma 1$\sim$3:
\begin{equation}
    I(X;Z)=I(X';Z)=I(X';Y)=I(U;Y)=I(U;V)
\end{equation}
\end{proof}
Therefore, the information of the whole squeezed channel is determined only by the markov chain $f$, i.e., the macro-dynamics. Thus $f$ is the bottleneck of the whole squeezed channel.

The Neural Information Squeezer framework can be converted as a Squeezed Information Channel as shown in Figure \ref{fig:squeezed_channel1}, on which, $\mathbf{x}_t,\mathbf{x}_t',\mathbf{y}_t,\mathbf{y}(t+1),\mathbf{y}(t+1)',\hat{\mathbf{x}}_{t+1}$, and $\mathbf{z}_{p-q}$ correspond to $X,X',U,V,Y,Z$, and $\mathcal{N}$, respectively. Therefore, 
\begin{equation}
\label{eq.bottleneck}
    I(\mathbf{y}_t;\mathbf{y}(t+1))=I(\mathbf{x}_t;\hat{\mathbf{x}}_{t+1})
\end{equation}

We can further extend Theorem \ref{thm.info_bottle} to the case of stacked neural information squeezer by the following corollary:

\begin{cor}
Theorem \ref{thm.info_bottle} can be extended to stacked neural information squeezer models.
\end{cor}
\begin{proof}
Because equation \ref{eq.bottleneck} holds for any neural information squeezer model whose encoder and decoder are composed by basic units of NIS, therefore, theorem \ref{thm.info_bottle} holds for stacked neural information squeezer models.
\end{proof}

\section{Proof of Theorem \ref{thm.training}}
\label{sec.training}
At first, we restate Theorem \ref{thm.training} as follow:

\textbf{Theorem \ref{thm.training}(Mutual information of the model will be closed to the data for a well trained framework)}: If the neural networks in NIS framework are well-trained, then:
\begin{equation}
    I(\hat{\mathbf{x}}_{t+1};\mathbf{x}_t)\approx I(\mathbf{x}_{t+1};\mathbf{x}_t)
\end{equation}

\begin{proof}
According to the objective function, i.e., Equation \ref{eq:loglikelihood} and \ref{eq:objectivefunction}, we know that if the neural networks in NIS framework are well-trained, that means the conditional probability distribution of the model will be closed to the one on the data, that is:
\begin{equation}
    p(\hat{\mathbf{x}}_{t+1}|\mathbf{x}_t)\simeq p(\mathbf{x}_{t+1}|\mathbf{x}_t)
\end{equation}
for all $\mathbf{x}_t\in \mathcal{R}^p$. Here, two distributions $p\simeq q$ means that $KL(p,q)\rightarrow 0$ as the training epoch is very large. Therefore:
\begin{equation}
    p(\mathbf{x}_t,\hat{\mathbf{x}}_{t+1})=p(\mathbf{x}_t)\cdot p(\hat{\mathbf{x}_{t+1}}|\mathbf{x}_t)\simeq p(\mathbf{x}_t)\cdot p(\mathbf{x}_{t+1}|\mathbf{x}_t)=p(\mathbf{x}_t,\mathbf{x}_{t+1}),
\end{equation}
and also:
\begin{equation}
    p(\hat{\mathbf{x}}_{t+1})=\int_{\mathbf{x}_t}p(\mathbf{x}_t,\hat{\mathbf{x}}_{t+1})d\mathbf{x}_t\simeq \int_{\mathbf{x}_t}p(\mathbf{x}_t,\mathbf{x}_{t+1})d\mathbf{x}_t=p(\mathbf{x}_{t+1}),
\end{equation}
so,
\begin{equation}
    I(\hat{\mathbf{x}}_{t+1};\mathbf{x}_t))\approx I(\mathbf{x}_{t+1};\mathbf{x}_t)
\end{equation}
\end{proof}

\begin{cor}
(The mutual information of macro-dynamics will not change if the model is well trained): For the well trained NIS model, the Mutual Information of the macro-dynamics $f_{\beta}$ will be irrelevant on all the parameters, including the scale $q$.
\end{cor}

\begin{proof}
According to Theorem \ref{thm.info_bottle} and Theorem \ref{thm.training}:
\begin{equation}
    I(\mathbf{y}_t;\mathbf{y}(t+1))=I(\mathbf{x}_t;\hat{\mathbf{x}}_{t+1})\approx I(\mathbf{x}_t;\mathbf{x}_{t+1}).
\end{equation}
This equality is irrelevant with all parameters in neural network and $q$.
\end{proof}


\section{Proof for Theorem \ref{thm.lowerbound}}
\label{sec.lowerbound}
\begin{lem}
\label{lemma.conditionalentropy}
For any continuous random variable $X$ and $Y$, we have:
\begin{equation}
I(X;Y)=H(Y)+\mathbb{E}_X(\ln|\det(\frac{\partial{Y}}{\partial{X}})|),
\end{equation}
where $\frac{\partial Y}{\partial X}$ is the Jacobian matrix. 
\end{lem}
\begin{proof}
According to the computation of the mutual information by a continuous mapping:
\begin{equation}
    I(X;Y)=H(Y)-H(Y|X),
\end{equation}
and:
\begin{equation}
    H(Y|X)=-\mathbb{E}_X(\ln|\det(\frac{\partial{Y}}{\partial{X}})|
\end{equation}
according to \cite{geiger2011}, thus:
\begin{equation}
I(X;Y)=H(Y)+\mathbb{E}_X(\ln|\det(\frac{\partial{Y}}{\partial{X}})|).
\end{equation}
\end{proof}

\textbf{Theorem \ref{thm.lowerbound}(Information on bottleneck is the lower bound of the encoder)}: For squeezed information chain shown in Figure \ref{fig:squeezed_channel}, the information of $U\rightarrow V$ is bounded by:
\begin{equation}
    I(U;V)\leq I(X;U)= H(U)+\mathbb{E}_X(\ln|\det(\frac{\partial{U}}{\partial{X}})|)\leq I(X;X')=H(X) + \mathbb{E}_X(\ln|\det(\frac{\partial{X'}}{\partial{X}})|)).
\end{equation}
\begin{proof}
Because both $\psi, \chi_q$, and $f$ are Markovian, so $X\rightarrow X'\rightarrow U\rightarrow V$ forms a Markov chain. Thus the data processing inequality holds:
\begin{equation}
    I(U;V)=I(X;Z)\leq I(X;U)\leq I(X;X'),
\end{equation}
and according to lemma \ref{lemma.conditionalentropy}:
\begin{equation}
    I(U;V)\leq H(U)+\mathbb{E}_X(\ln|\det(\frac{\partial{U}}{\partial{X}})|)\leq H(X) + \mathbb{E}_X(\ln|\det(\frac{\partial{X'}}{\partial{X}})|)).
\end{equation}
\end{proof}
Applying this theorem in the information squeezed channel (Figure \ref{fig:squeezed_channel}), we can obtain the form of Equation \ref{eq.lowerbound}.
\subsection{Proof for Theorem \ref{thm.ei_bijection}}
\label{sec.ei_calculation}
We re-state Theorem \ref{thm.ei_bijection}:

\textbf{Theorem \ref{thm.ei_bijection}(The mathematical expression for effective information of macro-dynamics)}: Suppose the probability density of $\mathbf{x}_{t+1}$ under given $\mathbf{x}_t$ can be described by a function $p(\mathbf{x}_{t+1}|\mathbf{x}_t)\equiv G(\mathbf{x}_{t+1},\mathbf{x}_t)$, and the Neural Information Squeezer framework is well trained, then the effective information of the macro-dynamics of $f_{\beta}$ can be calculated by:
\begin{equation}
    \label{eq:theorem_ei}
    EI_L(f_{\beta})=\frac{1}{(2L)^{p}}\cdot\int_{\sigma} \int_{\mathcal{R}^p} G(\mathbf{y},\psi_{\alpha}^{-1}(\mathbf{x}))\ln\frac{(2L)^{p}G(\mathbf{y},\psi_{\alpha}^{-1}(\mathbf{x}))}{\int_{\sigma}G(\mathbf{y},\psi_{\alpha}^{-1}(\mathbf{x}'))d\mathbf{x}'}d\mathbf{y} d\mathbf{x},
\end{equation}
where, $\sigma\equiv [-L,L]^p$ is the integration region for $\mathbf{x}$ and $\mathbf{x}'$.
\begin{proof}
According to Definition of effective information (EI),
\begin{equation}
    EI_L(f_{\phi_q})\equiv I\left(do\left(\mathbf{y}_t\equiv \mathbf{u}\sim U([-L,L]^q)\right);\mathbf{y}(t+1)\right).
\end{equation}
The effect of the do operator can be understood by another graphic model which is shown in \ref{fig:causal_graph}(a).
\begin{figure}
    \centering
    \includegraphics[scale=0.5]{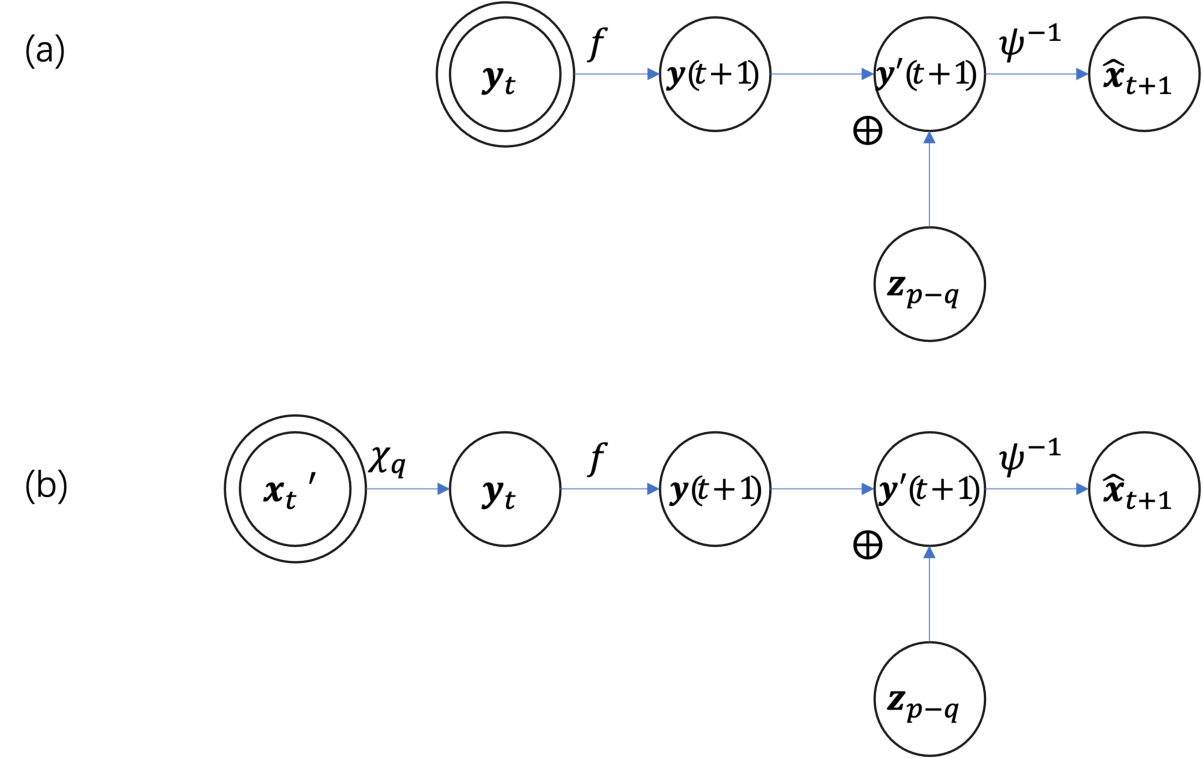}
    \caption{The graphic model of the squeezed information channel Figure \ref{fig:squeezed_channel1} after do operation. (a) do operator acts on the $\mathbf{y}_t$ node, and (b) do operator acts on the $\mathbf{x}_t'$ node. The nodes with double circles are the nodes that the do operator acts on.}
    \label{fig:causal_graph}
\end{figure}
For the squeezed information channel as shown in Figure \ref{fig:squeezed_channel1}, because $\mathbf{y}_t$ is the projection of $\mathbf{x}_t'$ on $q$ dimension, so if $\mathbf{x}_t'\sim U([-L,L]^p)$, then $\mathbf{y}_t\sim U([-L,L]^q)$, but the density increased by a factor $(2L)^{p-q}$. And because $\mathbf{x}_t\rightarrow \mathbf{x}_t'\rightarrow \mathbf{y}_t\rightarrow\mathbf{y}(t+1)$, So, the graphic model of Figure \ref{fig:causal_graph}(a) is equivalent to the graph in Figure \ref{fig:causal_graph}(b).
And according to Lemma \ref{lemma.projection} and Lemma \ref{lemma.concatenation}:
\begin{equation}
    I(\mathbf{y}_t;\mathbf{y}(t+1))=I(\mathbf{x}_t';\mathbf{y}(t+1)),=I(\mathbf{x}_t';\hat{\mathbf{x}}_{t+1})
\end{equation}
therefore,
\begin{equation}
\begin{aligned}
    EI_L(f_{\phi_q})&\equiv I\left(do\left(\mathbf{y}_t\sim U([-L,L]^q)\right);\mathbf{y}(t+1)\right)\\
    &=I\left(do(\mathbf{x}_t'\sim U([-L,L]^p));\mathbf{y}(t+1)\right)\\
    &=I\left(do(\mathbf{x}_t'\sim U([-L,L]^p));\hat{\mathbf{x}}_{t+1}\right).
\end{aligned}
\end{equation}
And,
\begin{equation}
\begin{aligned}
    &I\left(do(\mathbf{x}_t'\sim U([-L,L]^p));\hat{\mathbf{x}}_{t+1}\right)\\
    &=\int_{\sigma}\rho\cdot\left(\int_{\hat{\mathbf{x}}_{t+1}\in \mathcal{R}^q}p(\hat{\mathbf{x}}_{t+1}|\mathbf{x}_t')\ln\frac{p(\hat{\mathbf{x}}_{t+1}|\mathbf{x}_t')}{\rho\int_{\sigma} p(\hat{\mathbf{x}}_{t+1}|\mathbf{x}_t') d\mathbf{x}_t'}\cdot d\hat{\mathbf{x}}_{t+1}\right)d\mathbf{x}_{t}'\\
\end{aligned}
\end{equation}
where, $\sigma=[-L,L]^p$ is the integration region, and $\rho=(2L)^{-p}$. While the conditional probability of $\hat{\mathbf{x}}_{t+1}$ under given $\mathbf{x}_t$ is a function $p(\hat{\mathbf{x}}_{t+1}=y|\mathbf{x}_t=x)\equiv F(y,x)$, so:
\begin{equation}
    p(\hat{\mathbf{x}}_{t+1}|\mathbf{x}_t')=p(\hat{\mathbf{x}}_{t+1}|\mathbf{x}_t=\psi^{-1}(\mathbf{x}_t'))=F(\hat{\mathbf{x}}_{t+1},\psi^{-1}(\mathbf{x}_t'))
\end{equation} 
and according to Theorem \ref{thm.training}, if the NIS framework is well trained, we have:
\begin{equation}
    p(\hat{\mathbf{x}}_{t+1}|\mathbf{x}_t)=F(\hat{\mathbf{x}}_{t+1},\mathbf{x}_t)\approx G(\mathbf{x}_{t+1},\mathbf{x}_t)
\end{equation}
Therefore:
\begin{equation}
\begin{aligned}
    &I\left(do(\mathbf{x}_t'\sim U([-L,L]^p));\hat{\mathbf{x}}_{t+1}\right)\\
    &=\int_{\sigma}\rho\cdot\left(\int_{\hat{\mathbf{x}}_{t+1}\in \mathcal{R}^q}F(\hat{\mathbf{x}}_{t+1},\psi^{-1}(\mathbf{x}_t'))\ln\frac{F(\hat{\mathbf{x}}_{t+1},\psi^{-1}(\mathbf{x}_t'))}{\rho\int_{\sigma} F(\hat{\mathbf{x}}_{t+1},\psi^{-1}(\mathbf{x}_t')) d\mathbf{x}_t'}\cdot d\hat{\mathbf{x}}_{t+1}\right)d\mathbf{x}_{t}'\\
    &\approx \int_{\sigma}\rho\cdot\left(\int_{\mathbf{x}_{t+1}\in \mathcal{R}^q}G(\mathbf{x}_{t+1},\psi^{-1}(\mathbf{x}_t'))\ln\frac{G(\mathbf{x}_{t+1},\psi^{-1}(\mathbf{x}_t'))}{\rho\int_{\sigma} G(\mathbf{x}_{t+1},\psi^{-1}(\mathbf{x}_t')) d\mathbf{x}_t'}\cdot d\mathbf{x}_{t+1}\right)d\mathbf{x}_{t}'
\end{aligned}
\end{equation}
We then use $\mathbf{x}$ or $\mathbf{x}'$ to replace $\mathbf{x_t}'$ and use $\mathbf{y}$ to replace $\mathbf{x}_{t+1}$ in the integrations, then we have:
\begin{equation}
    EI_L(f_{\beta})\approx \frac{1}{(2L)^p}\int_{\sigma}\left(\int_{\mathbf{y}\in \mathcal{R}^q}G(\mathbf{y},\psi^{-1}(\mathbf{x}))\ln\frac{G(\mathbf{y},\psi^{-1}(\mathbf{x}))}{\rho\int_{\sigma} G(\mathbf{y},\psi^{-1}(\mathbf{x}')) d\mathbf{x}'}\cdot d\mathbf{y}\right)d\mathbf{x}
\end{equation}
\end{proof}

\section{Proof for Theorem \ref{thm.harder}}
\label{seq.harder}
\textbf{Theorem \ref{thm.harder}(Narrower is Harder)}: If $X$ is random variable with dimension $p$, and if the $q_1$ dimensional random variable $U_{q_1}$ is the projection of a $q_2$ dimensional variable $U_{q_2}$, and $0<q_1<q_2<p$, then:
\begin{equation}
    I(X;U_{q_1})\leq I(X;U_{q_2}).
\end{equation}

\begin{proof}
Because $q_1<q_2$, therefore, $U_{q_2}$ contains $U_{q_1}$ as the component, thus, there exists a $q_2-q_1$ dimensional random variable $U_{q_2-q_1}'$ such that:
\begin{equation}
    U_{q_2}=U_{q_1}\bigoplus U'_{q_2-q_1}
\end{equation}
Therefore:
\begin{equation}
    H(U_{q_2})=H(U_{q_1},U'_{q_2-q_1})=H(U_{q_1})+H(U'_{q_2-q_1}|U_{q_1})\geq H(U_{q_1})
\end{equation}
because $H(U'_{q_2-q_1}|U_{q_1})\geq 0$, and:
\begin{equation}
    \mathbb{E}_X(\ln|\det(\frac{\partial{U_{q_2}}}{\partial{X}})|\geq \mathbb{E}_X(\ln|\det(\frac{\partial{U_{q_1}}}{\partial{X}})|
\end{equation}
because the matrices of $\frac{\partial{U_{q_2}}}{\partial{X}}$ and $\frac{\partial{U_{q_2}}}{\partial{X}}$ are all sub-matrices of $\frac{\partial{\psi}}{\partial{X}}$ and the former contains the latter. Thus, according to lemma \ref{lemma.conditionalentropy}:
\begin{equation}
    I(X;U_{q_2})=H(U_{q_2})+\mathbb{E}_X(\ln|\det(\frac{\partial{U_{q_2}}}{\partial{X}})|\geq H(U_{q_1})+\mathbb{E}_X(\ln|\det(\frac{\partial{U_{q_1}}}{\partial{X}})=I(X;U_{q_1})
\end{equation}
\end{proof}
Thus, if the number of dimension is smaller, the mutual information between $X$ and $U$ will also be smaller. That means, narrower channel is harder to transfer information. 

Combined with Theorems \ref{thm.info_bottle} and \ref{thm.lowerbound}, we have the following inequalities for the squeezed information channel of Figure \ref{fig:squeezed_channel1}:
\begin{equation}
    I(\mathbf{x}_t;\hat{\mathbf{x}}_{t+1})\leq I(\mathbf{x}_t;\mathbf{y}_{t}^{q_1})\leq I(\mathbf{x}_t;\mathbf{y}_{t}^{q_2}).
\end{equation}
This is the form of Theorem \ref{thm.harder} in the main text.

\begin{adjustwidth}{-\extralength}{0cm}

\end{adjustwidth}
\bibliography{mybible.bib}
\end{document}